\documentclass{article}

\usepackage{arxiv}
\usepackage{times}
\usepackage{epsfig}
\usepackage{graphicx}
\usepackage{amsmath}
\usepackage{amssymb}
\usepackage{multirow}
\usepackage{color, colortbl}
\usepackage{algorithm}
\usepackage{algpseudocode}
\usepackage{wrapfig}
\usepackage{longtable}
\usepackage{array}

\newcommand{\PreserveBackslash}[1]{\let\temp=\\#1\let\\=\temp}
\newcolumntype{C}[1]{>{\PreserveBackslash\centering}m{#1}}
\newcolumntype{R}[1]{>{\PreserveBackslash\raggedleft}m{#1}}
\newcolumntype{L}[1]{>{\PreserveBackslash\raggedright}m{#1}}

\algdef{SE}[SUBALG]{Indent}{EndIndent}{}{\algorithmicend\ }%
\algtext*{Indent}
\algtext*{EndIndent}

\definecolor{Gray}{gray}{0.9}
\usepackage[pagebackref=true,breaklinks=true,colorlinks,bookmarks=false]{hyperref}

\author{
Mahsa Paknezhad\\
\textbf{Hamsawardhini Rengarajan}\\
\textbf{Chenghao Yuan}\\%\thanks{}
Bioinformatics Institute, A*STAR,\\
30 Biopolis Street, 07-01, Matrix\\
Singapore, 138671 \\
\texttt{mahsap@bii.a-star.edu.sg}\\
\And
Savitha Ramasamy\\
\textbf{Manas Gupta}\\
\textbf{Sujanya Suresh}\\
I2R, A*STAR,\\
1 Fusionopolis Way, 21-01 Connexis\\
Singapore 138632\\
\And
Hwee Kuan Lee\\
Bioinformatics Institute, A*STAR,\\
30 Biopolis Street, 07-01, Matrix\\
Singapore, 138671\\
National University of Singapore\\
Singapore, 119077\\
Singapore Eye Research Institute\\
20 College Road\\
Singapore, 169856\\
\texttt{leehk@bii.a-star.edu.sg}
}
\title{PaRT: Parallel Learning Towards Robust and Transparent AI}

\begin{document}

%%%%%%%%% TITLE
\maketitle
%%%%%%%%% ABSTRACT
\begin{abstract}
   This paper takes a parallel learning approach for robust and transparent AI. A deep neural network is trained in parallel on multiple tasks, where each task is trained only on a subset of the network resources. Each subset consists of network segments, that can be combined and shared across specific tasks. Tasks can share resources with other tasks, while having independent task-related network resources. Therefore, the trained network can share similar representations across various tasks, while also enabling independent task-related representations. The above allows for some crucial outcomes. (1) The parallel nature of our approach negates the issue of catastrophic forgetting. (2) The sharing of segments uses network resources more efficiently. (3) We show that the network does indeed use learned knowledge from some tasks in other tasks, through shared representations. (4) Through examination of individual task-related and shared representations, the model offers transparency in the network and in the relationships across tasks in a multi-task setting.  Evaluation of the proposed approach against complex competing approaches such as Continual Learning, Neural Architecture Search, and Multi-task learning shows that it is capable of learning robust representations. This is the first effort to train a DL model on multiple tasks in parallel. Our code is available at \href{https://github.com/MahsaPaknezhad/PaRT}{https://github.com/MahsaPaknezhad/PaRT}.
\end{abstract}

%%%%%%%%% BODY TEXT
%------------------------------------------------------------------------
\begin{figure}
    \centering
    \includegraphics[width=0.6\textwidth,trim={0.5cm, 0cm, 7cm, 0cm},clip]{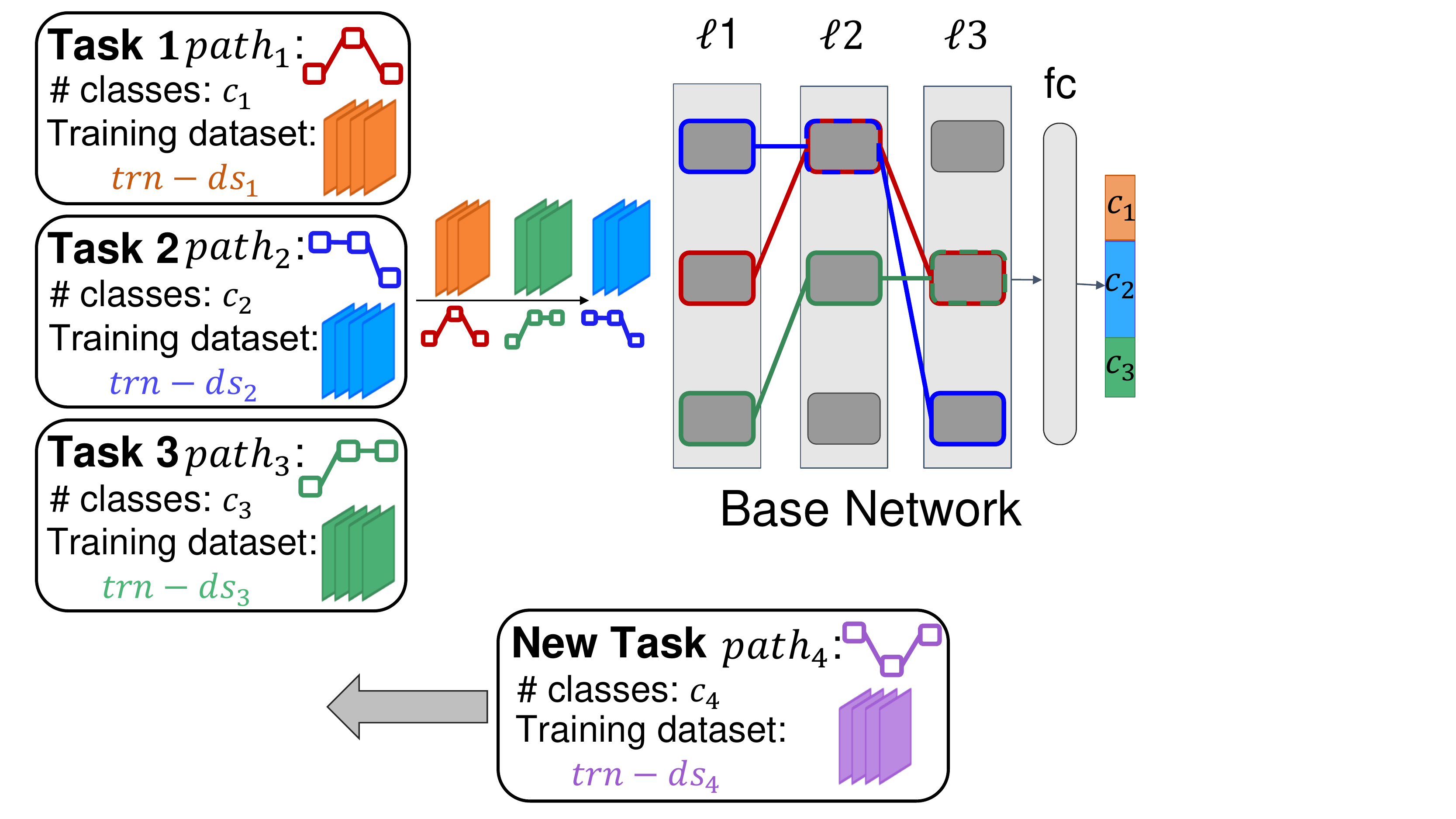}
    \caption{Figure shows a diagram of how parallel learning is carried out. Parallel learning encompasses a base neural network on which heterogeneous tasks are trained in parallel. Certain modules in the base network are assigned to each task and learning that task is done using those modules. Training on multiple tasks is performed by randomly sampling batches from the training dataset of those tasks. Once a new task is introduced, it will be added to the bucket of tasks on which the base network is trained.}
    \label{fig:parallel_alg}
\end{figure}
%----------------------------------------------------------------------
\section{Introduction}
Training deep learning (DL) models on multiple heterogeneous tasks is one of the main steps in the direction of offering robust and generalizable AI solutions. There exists an extensive body of research on multi-task learning \cite{zhang2021survey,zhang2018overview,thung2018brief}. However, the effectiveness of these methods in learning heterogeneous tasks with different data distributions is under question. Research in continual learning (CL) \cite{parisi2019continual,hadsell2020embracing} is recently gaining a lot of attention as it enables learning a sequence of tasks and leveraging on shared representations across those tasks while avoiding catastrophic forgetting of the previously learnt tasks. However, certain issues still remain in CL algorithms such as: 1) existing CL solutions have not been able to fully address catastrophic forgetting yet. 2) some CL solutions require more network resources to perform architectural adaptation, and 3) it is difficult to draw conclusions regarding representational similarities across different tasks with the current CL solutions. In this paper, we take a parallel approach to learning heterogeneous tasks using a single network. Parallel learning enables sharing representations across tasks, while also having task-related representations per individual tasks. Once a new task with a different input data distribution is introduced, the proposed algorithm can learn the new task alongside the existing tasks while sharing representations with the existing tasks. Parallel learning also draws a window of transparency through which we can see representational similarities as well as the amount of resource sharing in the network between tasks. CL is aimed at learning a sequence of tasks, while parallel learning is capable of learning multiple tasks in parallel. Furthermore, it can leverage on existing representations to learn new tasks swiftly. Parallel learning will be useful for learning multiple tasks from multiple sources of data within a single institution such as banks and hospitals. Parallel learning will equip these entities with a single DL model that learns from multiple datasets to produce multiple heterogeneous outcomes of interest with almost the same performance compared to a DL model trained on each dataset and outcome of interest. Our contributions are summarized below:
\begin{itemize}
\itemsep0.2em
    \item We develop an algorithm for training a DL model on multiple heterogeneous tasks with diverse input data distributions in parallel.
    \item We demonstrate that parallel learning is a simple but effective approach that has improved performance over multi-task learning, sequential CL algorithms, single task learning algorithms and the state-of-the-art algorithms in other domains. 
    \item Using our algorithm, we are able to see task-related and shared representations across tasks in a multi-task setting. We demonstrate that trained models using parallel learning learn more similar representations across tasks. 
\end{itemize}
%------------------------------------------------------------------------
\section{Related Work}
\textbf{Multi-task learning, Transfer learning and Meta-learning}: Multi-task learning \cite{long2015learning,zhou2021multi,ruder2017overview,vandenhende2020mti,misra2016cross} and transfer learning \cite{tan2018survey,zhuang2020comprehensive} algorithms aim to reduce the required amount of data and the need for different models by using a single model to learn different tasks. Multilinear Relationship Network (MRN) \cite{long2015learning} is one of the recently proposed multi-task learning algorithms. MRN jointly learns transferable features and multilinear relationships of tasks and features. It is important to note that the parallel learning is not a multi-task learning algorithm. In multi-task learning, a DL model is trained to produce multiple outcomes of interest on an input dataset. In parallel learning, however, different outcomes of interest are defined on different input datasets. Also, our proposed DL model is different from transfer learning. Parallel learning shares a portion of the neurons across multiple tasks and trains them on those tasks simultaneously whereas in transfer learning, neurons are initialized with values learned on an earlier task and are trained on a new task. Another group of algorithms called Meta-learning \cite{liu2020distribution,hospedales2020meta,rusu2018meta} aim to reduce the required amount of data to learn a new task using trained models on other tasks. Although meta-learning algorithms have been proposed to learn tasks with heterogeneous attribute spaces \cite{iwata2020meta}, these methods are yet to be extended to heterogeneous tasks. As will be shown in section \ref{sec:results}, parallel learning can be used to learn heterogeneous tasks with diverse input data distributions. \par
%----------------------------------------------------------------------
\begin{figure*}
    \centering
    \includegraphics[width=\textwidth, trim={0cm 8cm 0cm 0cm}, clip]{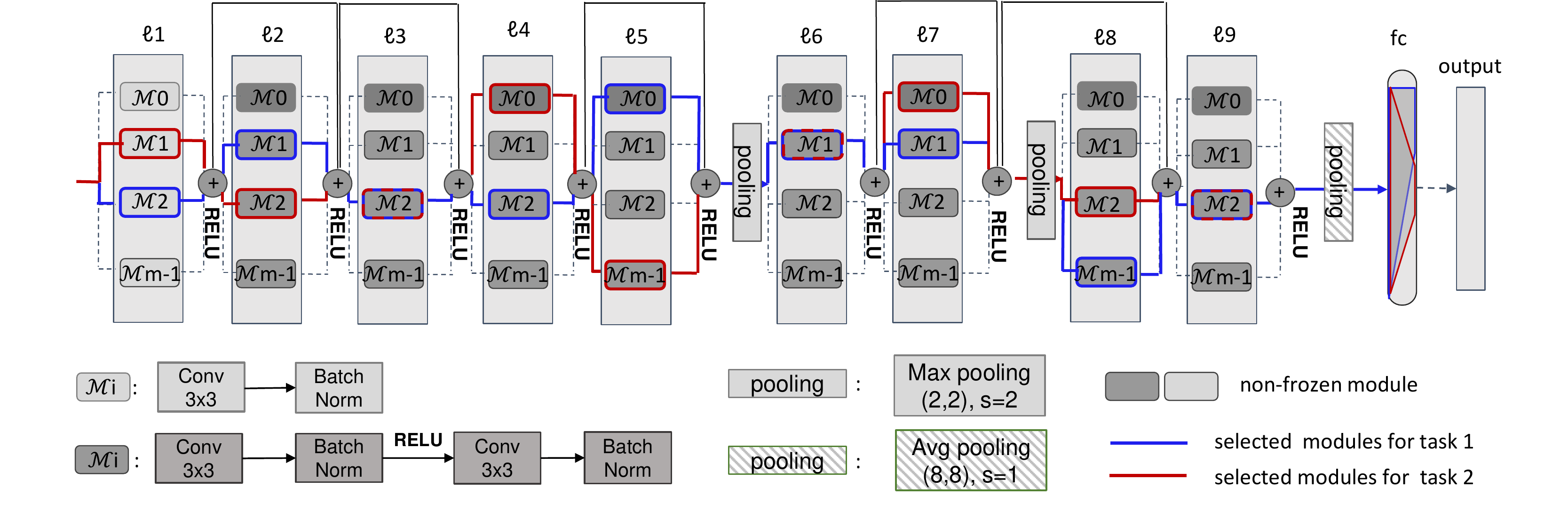}
    \caption{Figure shows a simple architecture for the base network in our parallel learning algorithm. One row of modules is equal to a ResNet-18 architecture in this figure. For each task, a set of $N$ modules is randomly chosen in each layer and are assigned to the task. Training on the task is performed only on the selected modules for that task.}
    \label{fig:base_network}
\end{figure*}
%----------------------------------------------------------------------
\textbf{Neural architecture search (NAS)}: Studies in the NAS domain \cite{isensee2018nnu,zoph2018learning,zoph2016neural,kirsch2018modular,gou2020clearer,negrinho2017deeparchitect,liu2018progressive,yao2020sm,real2017large} target automatic and fast design of DL models for the task in hand. pDarts \cite{chen2019progressive} is an optimized version of the Darts algorithm \cite{liu2018darts} which is one of the most benchmarked algorithms in the NAS domain. In pDarts, a network is formed by stacking multiple cells together. The architecture within a cell is derived using a gradient-based approach. ENAS \cite{pham2018efficient} is another algorithm which trains a controller using policy gradient to search an optimal network architecture. A multi-agent approach to NAS where agents control parts of the network and coordinate to reach the optimal architectures is proposed as MANAS \cite{carlucci2019manas}. However, NAS algorithms generate a separate DL model for each task. Consequently, while they reduce the time needed to develop a model, they are not considered to be memory-efficient and they do not take advantage of learned representations from other tasks.\par
\textbf{Continual learning (CL)}: A large amount of work has been done in CL \cite{lesort2020continual,kirkpatrick2017overcoming,shin2017continual,rajasegaran2019random,hung2019compacting} to train a DL model on new tasks while retaining its performance on preceding tasks. Training on tasks is done sequentially in many such algorithms. Once training on a task finishes, the contributing neurons to that task are frozen or are allowed to change slightly to avoid catastrophic forgetting. In Elastic Weight Consolidation (EWC) \cite{kirkpatrick2017overcoming}, for instance, a regularization term pulls back parameter values to their old learned values on previous tasks. Hung et al. \cite{hung2019compacting} train multiple tasks sequentially by expanding the neural network if necessary. They add a model compression step before training the DL model on a new task. A closer group of algorithms to our proposed method are CL algorithms with generative replay algorithms such as the work by Shin et al. \cite{shin2017continual}, van de Ven et al. \cite{van2020brain} and other methods \cite{van2018generative,lesort2019generative,aljundi2019online} where a generator is trained to synthesize samples for previous task, or a small representative set of data from previously learned tasks called a coreset \cite{lopez2017gradient,rebuffi2017icarl,lopez2017gradient} is kept. The samples are mixed with the training data of the new task and are used during training. Random Path Selection (RPS) \cite{rajasegaran2019random}, is a good example which deploys coresets. The main reason for training a generator or keeping coresets is to make the algorithm more memory-efficient. We, however, target entities such as hospitals at which clinical data will always be kept. As a result, memory-efficiency is not a major concern. What is important, instead, is to train a DL model which can learn to do multiple important tasks such as cancer grading and disease prediction with high accuracy and limited amount of data, which is the main objective of the proposed approach in this paper. 
%------------------------------------------------------------------------
\section{Method}
In this section, we introduce PaRT: a parallel learning algorithm for robust and transparent AI. This is the first effort to train a DL model on multiple tasks in parallel. We draw inspirations for our algorithm from existing efforts in the CL domain. PaRT encompasses an algorithm for parallel learning of multiple tasks using a base network architecture. A simple diagram of our algorithm is show in Fig. \ref{fig:parallel_alg}. In the following, we explain our proposed algorithm. \par
%----------------------------------------------------------------------

\begin{algorithm}[h]
\caption{Parallel Learning}\label{alg:parallel}
\hspace*{\algorithmicindent}\textbf{input}: $k$ : The number of tasks\\
\hspace*{\algorithmicindent}($trn\text{-}ds_i$, $val\text{-}ds_i$): The training and validation datasets for the $i^{th}$ task\\
\hspace*{\algorithmicindent}$c_i$: The number of classes for the $i^{th}$ classification task\\
\hspace*{\algorithmicindent}$C_\theta$: The base network\\
\hspace*{\algorithmicindent}$path_i$: A randomly selected set of modules for the $i^{th}$ task\\
 \hspace*{\algorithmicindent}\textbf{output}: $C_\theta$ trained on the $k$ tasks
\begin{algorithmic}[1]
\Procedure{ParallelLearning}{}
\State \{$trn\text{-}ds^{new}_1$, $trn\text{-}ds^{new}_2$, ..., $trn\text{-}ds^{new}_k$\} $\xleftarrow{}$ Update the $trn\text{-}ds_i$ for $i=1,...,k$ to have equal size by oversampling
\State \textbf{for each} training epoch
\State \hskip1.4em \textbf{while} there exists untrained batches
\State \hskip2.7em $bucket \xleftarrow{} \{trn\text{-}ds^{new}_1, ... , trn\text{-}ds^{new}_k\}$
\State  \hskip2.7em Randomly sample a dataset $trn\text{-}ds^{new}_i$  from the $bucket$ without replacement
\State \hskip2.7em Sample a few batches $batch\text{-}set$ from  $trn\text{-}ds^{new}_i$
\State  \hskip2.7em Set $start = \sum_{n=1}^{i-1}c_n$ and $end = \sum_{n=1}^{i}c_n$
\State  \hskip2.7em \textbf{for} $batch_j$ from $batch\text{-}set$
\State  \hskip4em Train $C_\theta$ on $batch_j$ over $path_i$ using output neurons $[start,end]$ 
\EndProcedure
\end{algorithmic}
\end{algorithm}
%----------------------------------------------------------------------
Fig. \ref{fig:base_network} shows the architecture we deployed as our base network. The base network consists of multiple layers, with a total of $M$ modules in each layer. One row in the base network can be equivalent to any standard network architecture such as ResNet-18 or ResNet-50 architectures. Let us assume there are $k$ tasks to learn using the proposed parallel learning algorithm and the base network. At the beginning, the training datasets for all tasks are oversampled to have the same size. For each task, $N$ modules will be randomly selected from the existing $M$ modules in each layer independent of the other layers and are assigned to the task for training. The parameters $N$, $N$ are hyperparameters. Some modules may be selected by multiple tasks. As a result, such modules will be shared and trained on those tasks. Let us define the randomly selected set of modules at all the layers in the base network for the $i^{th}$ task as $path_i$. The training process on the base network is as follows: In each training epoch, a few batches called $batch\text{-}set$s are taken from all k training datasets. We then trained these $batch\text{-}set$s in a random order. Training on batches from $task i$ is performed on the modules specified in $path_i$ in the base network. We then repeated the process until all samples have been taken from the training sets and used for training, and one epoch finishes. The size of the $batch\text{-}set$ is also a hyperparameter. The outputs of modules selected in the same layer for task i are summed together, and the sum is fed to the modules selected by task i in the next layer(s). No module is frozen during training, and tasks are trained on their corresponding modules. The number of neurons in the output layer is the sum of the number of classes for all tasks. The parallel nature of training in the proposed algorithm negates catastrophic forgetting. We use $ADAM$ as our optimization  and $Cross Entropy$ for our loss function. Our parallel learning algorithm for training the base network on these $k$ tasks is shown in Algorithm \ref{alg:parallel}. Our code will be available online. \par
Once training the base network finishes, validation for the $i^{th}$ task is performed by passing samples from the validation dataset of the task, $val\text{-}ds_i$, through $path_i$ and comparing the predicted labels with ground truth labels. We compare our algorithm to sequential learning, a basic version of Continual Learning. In sequential learning, the model is trained on tasks one after another. Similar to parallel learning, a set of modules defined by $path_i$ are assigned to each task $i$. Once training for task $i$ finishes, the modules in $path_i$ will be frozen and will no longer be trainable for subsequent tasks even if those frozen modules are assigned to a new task. We also benchmark our algorithm against single task learning which is defined as training $path_i$ in the base network from scratch on only task $i$. Single task learning represents the achievable performance for each task on our base network. In the next section, we will apply our parallel learning algorithm to learning multiple homogeneous as well as heterogeneous tasks. We will see certain changes have to be made to our base network architecture when Parallel Learning is applied on heterogeneous tasks.\par
We use the representation similarity metric proposed by Kornblith et al. \cite{kornblith2019similarity} to compare the learned task-related and shared representations by models trained using PaRT with the learned representations by models trained on each task using single task learning. This metric is called centered kernel alignment (CKA) and is closely related to canonical correlation analysis (CCA). CKA is a stronger metric compared to CCA and other similarity metrics as it allows measuring meaningful similarities between representations of higher dimension than the number of data points.
%----------------------------------------------------------------------
\section{Experimental Evaluation and Analysis}
\label{sec:results}
We designed three experiments with increasing heterogeneity in tasks. In the following, we will explain each experiment and analyze the results for the experiments. 
\subsection{Training on Homogeneous Tasks}
\label{sec:homo_cifar100}
\begin{figure}[!h]  
    \centering
    \begin{tabular}{ccc}
    \includegraphics[width=0.25\textwidth, trim={0.8cm 0.5cm 0.5cm 0.5cm}, clip]{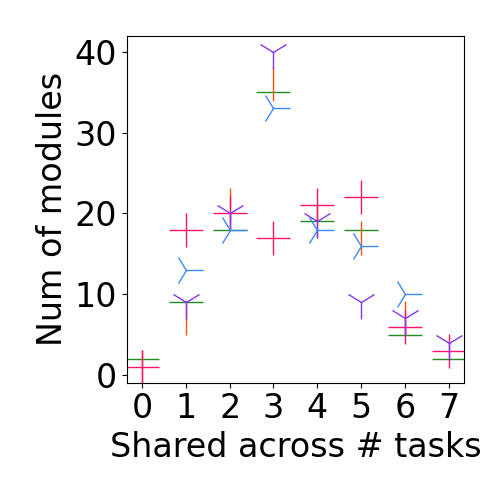} &
    \includegraphics[width=0.25\columnwidth, trim={0.8cm 0.5cm 0.5cm 0.5cm}, clip]{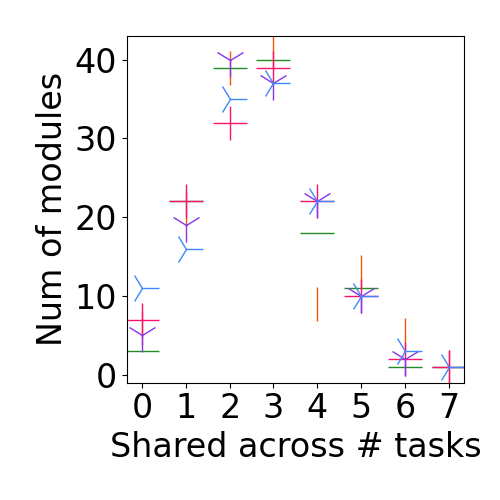} &
    \includegraphics[width=0.25\columnwidth, trim={0.8cm 0.5cm 0.5cm 0.5cm}, clip]{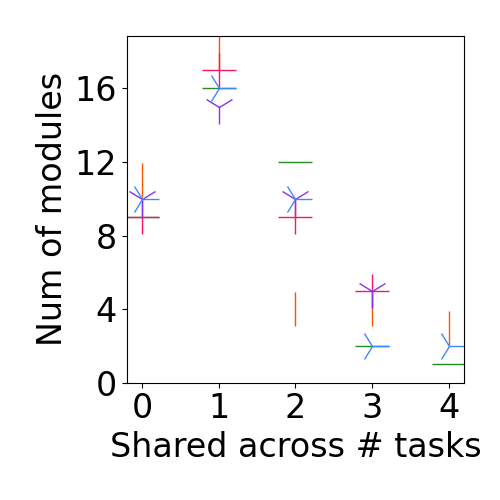}\\
    \end{tabular}
    \caption{Figure shows the module sharing profile when using PaRT for the CIFAR100 experiments (Left), the CIFAR10 and CIFAR100 experiments (Centre) and FiveTasks experiments (Right).}
    \label{fig:modulesharing}
\end{figure}

In this section, we applied PaRT on 10 homogeneous tasks defined on the CIFAR100 \cite{krizhevsky2009learning} dataset. Each task consists of classifying images from 10 randomly selected classes from the CIFAR100 dataset into 10 categories. ResNet-18 network was used to generate the base network. The total number of modules in each layer in the base network was 12 ($M$=12). Four modules per layer were randomly selected and assigned to each task ($N$=4). Due to random assignment of modules to different tasks, certain modules were shared between different tasks. The module sharing profile for this experiment is shown in Fig. \ref{fig:modulesharing} (Left). As an example, the plot shows that around 20 modules were shared among four tasks for PaRT experiments. \par
\begin{figure*}
    \centering
    \resizebox{\textwidth}{!}{
    \begin{tabular}{cc}
            \includegraphics[width=0.48\textwidth, trim={0.8cm 0cm 0.7cm 0.5cm},clip]{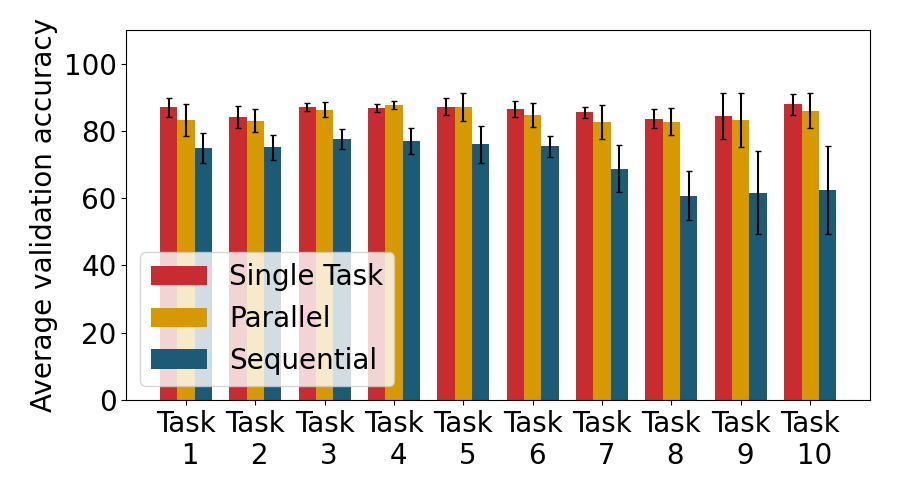}  &  \includegraphics[width=0.48\textwidth, trim={0.8cm 0cm 0.7cm 0.5cm},clip]{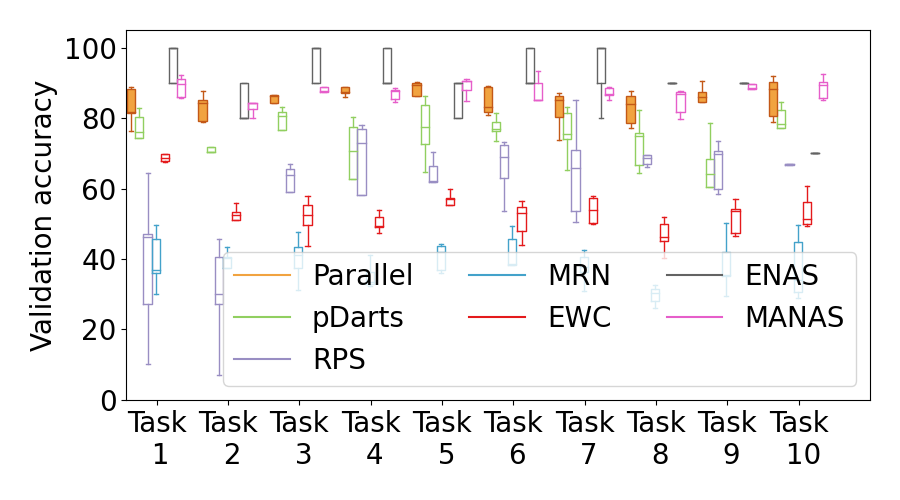} \\
    \end{tabular}}
    \caption{Mean validation accuracy of DL models trained on 10 tasks defined on the CIFAR100 \cite{krizhevsky2009learning} dataset. Each task consists of classifying 10 randomly selected classes in the CIFAR100 dataset into 10 categories. (Left) compares the performance of PaRT with single task learning and sequential learning methods. (Right) shows performance of PaRT and those of the competing algorithms. Five DL models were trained for each algorithm.}
    \label{fig:cifar100}
\end{figure*}

In five runs using five different seeds, tasks were defined as explained above and certain modules were assigned to each task randomly. The selected modules were trained on their corresponding task following three different learning methods: 1) single task learning, 2) sequential learning and 3) PaRT. For all learning methods, training was performed until convergence. Fig. \ref{fig:cifar100} (Left) compares the validation accuracy of models trained using these three learning methods. One can see that the average validation accuracy of DL models trained with single task learning method is high for the learned task but very low for the other nine unlearned tasks. Validation accuracy of the DL models trained using PaRT is closer to the achievable performance for each task compared to using sequential learning. It can also be seen that the validation accuracy of the second half of the tasks for sequential learning gradually decreases. We believe the reason is that the base network runs out of unfrozen modules to train on these tasks, and hence insufficient representations are learned for these tasks. \par
%----------------------------------------------------------------------
%----------------------------------------------------------------------
\begin{table*}[]
    \centering
    \resizebox{\textwidth}{!}{%
    \begin{tabular}{|lccccccccccc|}
    \rowcolor{Gray}
   \hline
   {Learning Method} & Task 1 & Task 2 & Task 3 & Task 4 & Task 5 & Task 6 & Task 7 & Task 8 & Task 9 & Task 10 & {Avg}\\
    \hline
    \multirow{5}{*}{Single Task} & 86.96 & 84.12 & 87.04 & 86.88 & 87.24 & 86.38 & 85.58 & 83.68 & 84.52 & 87.85 & \multirow{2}{*}{86.03} \\
     & \footnotesize{\textpm~2.89} & \footnotesize{\textpm~3.23}& \footnotesize{\textpm~1.11}&  \footnotesize{\textpm~1.23}&  \footnotesize{\textpm~2.64}&  \footnotesize{\textpm~2.37}&  \footnotesize{\textpm~1.61} & \footnotesize{\textpm~2.95}&  \footnotesize{\textpm~6.85}&  \footnotesize{\textpm~3.13}& \\
     \cline{2-12}
     & \cellcolor{Gray}Tasks & \cellcolor{Gray}Tasks & \cellcolor{Gray}Tasks & \cellcolor{Gray}Tasks & \cellcolor{Gray}Tasks & \cellcolor{Gray}Tasks & \cellcolor{Gray}Tasks & \cellcolor{Gray}Tasks & \cellcolor{Gray}Tasks & \cellcolor{Gray}Tasks& \cellcolor{Gray}\\
     & \cellcolor{Gray}$All \backslash {1}$ & \cellcolor{Gray}$All \backslash {2}$& \cellcolor{Gray}$All \backslash {3}$& \cellcolor{Gray}$All \backslash {4}$& \cellcolor{Gray}$All \backslash {5}$& \cellcolor{Gray}$All \backslash {6}$& \cellcolor{Gray}$All \backslash {7}$& \cellcolor{Gray}$All \backslash {8}$& \cellcolor{Gray}$All \backslash {9}$& \cellcolor{Gray}$All \backslash {10}$& \cellcolor{Gray}\\
     \cline{2-12}
     & 10.11 & 9.58 & 10.03 & 9.36 & 10.36 & 10.06 & 10.01 & 9.39 & 9.5 & 9.89 & 9.83\\
     & \footnotesize{\textpm1.58} & \footnotesize{\textpm1.53} & \footnotesize{\textpm1.31} & \footnotesize{\textpm0.82} & \footnotesize{\textpm0.83} & \footnotesize{\textpm1.31} & \footnotesize{\textpm1.29} & \footnotesize{\textpm1.29} & \footnotesize{\textpm1.66} & \footnotesize{\textpm3.49} & \\
    \rowcolor{Gray}
   \hline
   {Learning Method} & Task 1 & Task 2 & Task 3 & Task 4 & Task 5 & Task 6 & Task 7 & Task 8 & Task 9 & Task 10 & \\
    \hline
    \multirow{2}{*}{Sequential} & 74.96  & 75.06  & 77.48 & 76.96 & 76.02  & 75.40  & 68.80 & 60.76 & 61.66 & 62.42 & \multirow{2}{*}{70.95}\\
     & \footnotesize{\textpm~4.54} & \footnotesize{\textpm~3.66} & \footnotesize{\textpm~3.01} &  \footnotesize{\textpm~3.86} & \footnotesize{\textpm~5.49} & \footnotesize{\textpm~3.04} & \footnotesize{\textpm~7.10} & \footnotesize{\textpm~7.32}& \footnotesize{\textpm~12.41} & \footnotesize{\textpm~13.23} & \\
    \hline
    \multirow{2}{*}{Parallel} & 80.82 & 75.96 & 79.08 & 73.06 & 75.64 & 78.44 & 79.06 & 80.30 & 84.46 & 89.66 & \multirow{2}{*}{79.64} \\  
     &\footnotesize{\textpm~5.16} & \footnotesize{\textpm~6.34} & \footnotesize{\textpm~3.26} & \footnotesize{\textpm~6.84} & \footnotesize{\textpm~9.38} & \footnotesize{\textpm~5.25} & \footnotesize{\textpm~2.99} & \footnotesize{\textpm~5.27} & \footnotesize{\textpm~3.84} & \footnotesize{\textpm~3.51} & \\  
    \hline
    \end{tabular}}
    \caption{Table compares validation accuracy of models trained on 10 different tasks defined on the CIFAR100 dataset for single task learning, sequential learning and PaRT. For DL models trained using single task learning, the measured validation accuracy on the learned task and unlearned tasks are provided. Five different DL models were trained for each task and learning method and the validation accuracy of the five models were averaged. }
    \label{tab:cifar100}
\end{table*}
%----------------------------------------------------------------------
Fig. \ref{fig:cifar100} (Right) compares accuracy of DL models trained on the defined 10 tasks using PaRT with competing algorithms in CL such as RPS \cite{rajasegaran2019random}  and EWC \cite{kirkpatrick2017overcoming}, NAS such as pDarts  \cite{chen2019progressive}, ENAS \cite{pham2018efficient} and MANAS \cite{carlucci2019manas}, and multi-task learning such as MRN \cite{long2015learning}. The experiment setup and hyperparameter values for our method and competing algorithms are provided in Appendix B and the mean and standard deviation of the validation accuracy for each competing algorithm is provided in Appendix C of the Supplementary Materials. It is shown that parallel learning using the simple base network architecture shown in Fig. \ref{fig:base_network} and with random assignment of modules can outperform many existing complex algorithms from different DL domains. Only a NAS algorithm, MANAS, constantly outperformed PaRT in this experiment. Table \ref{tab:cifar100} shows the mean and standard deviation of the measured validation accuracy for the trained models for PaRT, sequential and single task learning methods. 

%----------------------------------------------------------------------
%----------------------------------------------------------------------
\begin{figure*}
    \centering
    \resizebox{\textwidth}{!}{
    \begin{tabular}{cc}
        \includegraphics[width=0.48\textwidth, trim={0.8cm 0cm 0.7cm 0.5cm},clip]{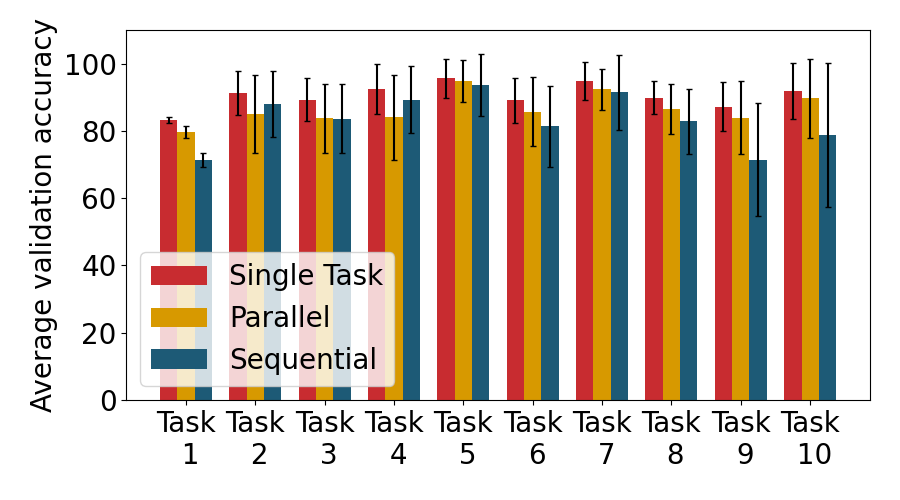}  &  \includegraphics[width=0.48\textwidth, trim={0.8cm 0cm 0.7cm 0.5cm},clip]{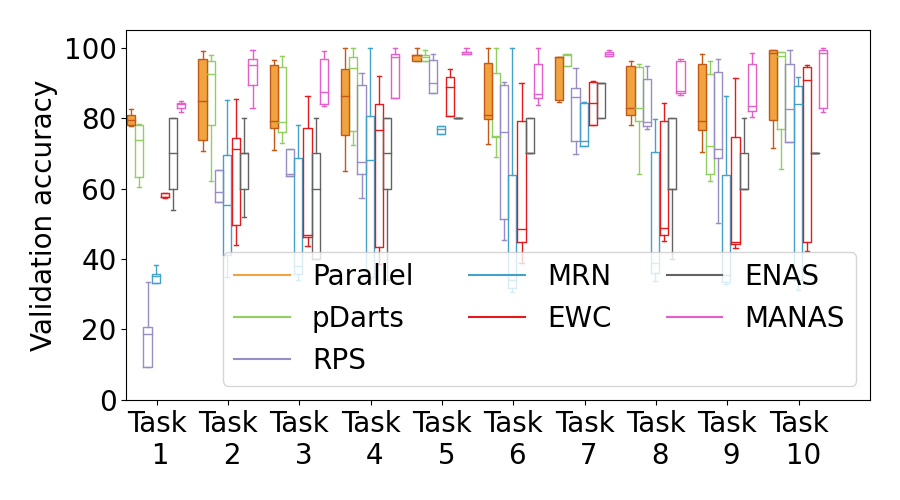} \\
    \end{tabular}}
    \caption{Mean validation accuracy of DL models trained on five tasks defined on the CIFAR10 dataset and five tasks defined on the CIFAR100 \cite{krizhevsky2009learning} dataset. The tasks defined on the CIFAR10 dataset classify two randomly selected classes from the CIFAR10 dataset into two categories. The tasks defined on the CIFAR100 dataset classify 20 randomly selected classes from the CIFAR100 dataset into 20 categories. (Left) compares the performance of PaRT with single task learning and sequential learning. (Right) compares the performance of PaRT with those of the competing algorithms. Five models were trained for each learning method. }
    \label{fig:cifar100_10}
\end{figure*}
%----------------------------------------------------------------------
\begin{table*}[t]
    \centering
    \resizebox{\textwidth}{!}{%
    \begin{tabular}{|lccccccccccc|}
    \rowcolor{Gray}
     \hline
    {Learning Method} & Task 1 & Task 2 & Task 3 & Task 4 & Task 5 & Task 6 & Task 7 & Task 8 & Task 9 & Task 10 & {Avg}\\
    \hline
    \multirow{5}{*}{Single Task}&  83.29&  91.30& 89.24 & 92.45
 & 95.64 & 89.07&  94.90& 89.82 & 87.22& 91.85& \multirow{2}{*}{90.48}\\
     & \footnotesize{\textpm~0.96} & \footnotesize{\textpm~6.48}& \footnotesize{\textpm~6.45}&  \footnotesize{\textpm~7.57}&  \footnotesize{\textpm~5.75}&  \footnotesize{\textpm~6.62}&  \footnotesize{\textpm~5.63} & \footnotesize{\textpm~4.91}&  \footnotesize{\textpm~7.39}&  \footnotesize{\textpm~8.34}&  \\
     \cline{2-12}
     & \cellcolor{Gray}Tasks & \cellcolor{Gray}Tasks & \cellcolor{Gray}Tasks & \cellcolor{Gray}Tasks & \cellcolor{Gray}Tasks & \cellcolor{Gray}Tasks & \cellcolor{Gray}Tasks & \cellcolor{Gray}Tasks & \cellcolor{Gray}Tasks & \cellcolor{Gray}Tasks &\cellcolor{Gray} \\
    & \cellcolor{Gray}$All \backslash {1}$ & \cellcolor{Gray}$All \backslash {2}$& \cellcolor{Gray}$All \backslash {3}$& \cellcolor{Gray}$All \backslash {4}$& \cellcolor{Gray}$All \backslash {5}$& \cellcolor{Gray}$All \backslash {6}$& \cellcolor{Gray}$All \backslash {7}$& \cellcolor{Gray}$All \backslash {8}$& \cellcolor{Gray}$All \backslash {9}$& \cellcolor{Gray}$All \backslash {10}$& \cellcolor{Gray}\\ 
    \cline{2-12}
    & 28.64 & 27.55 & 26.44 & 27.31 & 24.74 & 25.16 & 26.95 & 25.70 & 24.87 & 25.16 & 26.25 \\
    & \footnotesize{\textpm~8.11} & \footnotesize{\textpm~8.87} & \footnotesize{\textpm~7.57} & \footnotesize{\textpm~8.27} & \footnotesize{\textpm~8.55} & \footnotesize{\textpm~10.97} & \footnotesize{\textpm~2.19} & \footnotesize{\textpm~7.71} & \footnotesize{\textpm~7.13} & \footnotesize{\textpm~7.35} &\\
    \hline
    \rowcolor{Gray}
    Learning Method &Task 1 & Task 2 & Task 3 & Task 4 & Task 5 & Task 6 & Task 7 & Task 8 & Task 9 & Task 10 &\\
    \hline
   \multirow{2}{*}{Sequential} & 71.28 & 88.05
 & 83.62 & 89.32 & 93.54 & 81.33 & 91.51 & 82.84 & 71.47 &78.71 & \multirow{2}{*}{83.17}\\
     & \footnotesize{\textpm~2.11} & \footnotesize{\textpm~9.81}& \footnotesize{\textpm~10.24}&  \footnotesize{\textpm~10.10}&  \footnotesize{\textpm~9.21}&  \footnotesize{\textpm~11.97}&  \footnotesize{\textpm~11.11} & \footnotesize{\textpm~9.67}&  \footnotesize{\textpm~16.80}&  \footnotesize{\textpm~21.47}&  \\
    \hline
    \multirow{2}{*}{Parallel} & 79.75 & 85.00
 & 83.77 & 84.10 & 94.88 & 85.76 & 92.33 & 86.61 & 83.94 & 89.63 & \multirow{2}{*}{86.58}\\
& \footnotesize{\textpm~1.74} & \footnotesize{\textpm~11.58}& \footnotesize{\textpm~10.22}&  \footnotesize{\textpm~12.62}&  \footnotesize{\textpm~6.17}&  \footnotesize{\textpm~10.37}&  \footnotesize{\textpm~6.15} & \footnotesize{\textpm~7.45}&  \footnotesize{\textpm~10.89}&  \footnotesize{\textpm~11.85}&  \\
    \hline
    \end{tabular}}
    \caption{Table compares validation accuracy of DL models trained on 10 different tasks defined on the CIFAR10 and CIFAR100 datasets for single task learning, sequential learning and PaRT. For DL models trained using single task learning, the measured validation accuracy on the learned task and unlearned tasks are provided. Five different models were trained for each task and learning method and the validation accuracy of the five models were averaged.}
    \label{tab:cifar100_10}
\end{table*}
%----------------------------------------------------------------------
\subsection{Heterogenous Tasks}
In this section, we applied PaRT on more heterogeneous tasks. Two experiments were conducted. The experiments are presented in increasing order of complexity.
\vspace{-7pt}
\subsubsection{CIFAR10 and CIFAR100 Experiment}
\label{sec:hetero_cifar10_cifar100}
In this experiment, five tasks were defined on the CIFAR10 \cite{krizhevsky2009learning} dataset, and five tasks were defined on the CIFAR100 \cite{krizhevsky2009learning} dataset. The tasks defined on the CIFAR10 dataset are about classifying images from two randomly selected classes in the CIFAR10 dataset into two categories; the tasks defined on the CIFAR100 dataset are about classifying images from 20 randomly selected classes in the CIFAR100 dataset into 20 categories. ResNet-18 network was used to generate the base network. The total number of modules in each layer in the base network was 15 ($M$=15). Four modules per layer were randomly selected and assigned to each task ($N$=4). The module sharing profile for this experiment is shown in Fig. \ref{fig:modulesharing} (Centre). \par
Five DL models were trained for each learning method. Fig. \ref{fig:cifar100_10} (Left) shows the results for this experiment using PaRT, single task learning, and sequential learning.  Table \ref{tab:cifar100_10} shows the mean and standard deviation of the measured validation accuracy for the trained models for PaRT, sequential and single task learning methods. Similar to the previous experiment, the mean validation accuracy for DL models trained using single task learning on the learned tasks is higher that PaRT and sequential learning. However, their performance on the other nine tasks is found to be low. In addition, even though we increased the total number of modules in each layer ($M$) to 15, the DL models trained using the Sequential Learning method still do not have as good performance on the defined 10 tasks as the DL models trained using PaRT. Similar behaviour is seen when comparing the validation accuracy of the DL models trained on the 10 tasks using PaRT with competing algorithms (Fig. \ref{fig:cifar100_10} (Right)). The experiment setup and hyperparameter values for our method and each competing algorithm is provided in Appendix B and the mean and standard deviation of the validation accuracy for each competing algorithm is provided in Appendix C of the Supplementary Materials. 
%----------------------------------------------------------------------
\begin{table}[b]
    \centering
     \resizebox{0.5\columnwidth}{!}{%
    \begin{tabular}{|llccc|}
        \rowcolor{Gray}
         \hline
        Task & Dataset & \#Train & \#Eval & \#Classes\\
        \hline
        Task 1 & CUBs \cite{wah2011caltech} & 5,994 & 5,794 & 200\\
        \hline
        Task 2 & Stanford Cars \cite{krause20133d}& 8,144 & 8,041 & 196\\
        \hline
        Task 3 & Flowers \cite{nilsback2008automated}& 2,040 & 6,149 & 102\\
        \hline
        Task 4 & WikiArt \cite{saleh2015large} & 42,129 & 10,628 & 195\\
        \hline
        Task 5 & Sketch \cite{eitz2012hdhso} & 16,000 & 4,000 & 250\\
        \hline
    \end{tabular}}
    \caption{Table shows brief information about the five datasets that were deployed to define five heterogeneous classification tasks in section \ref{sec:hetero_5tasks}.}
    \label{tab:datasets}
\end{table}
%----------------------------------------------------------------------
\subsubsection{FiveTasks Experiment}
\label{sec:hetero_5tasks}
%----------------------------------------------------------------------
\begin{figure}[t]
    \centering
    \includegraphics[width=0.6\columnwidth, trim={4cm 5cm 14cm 3cm},clip]{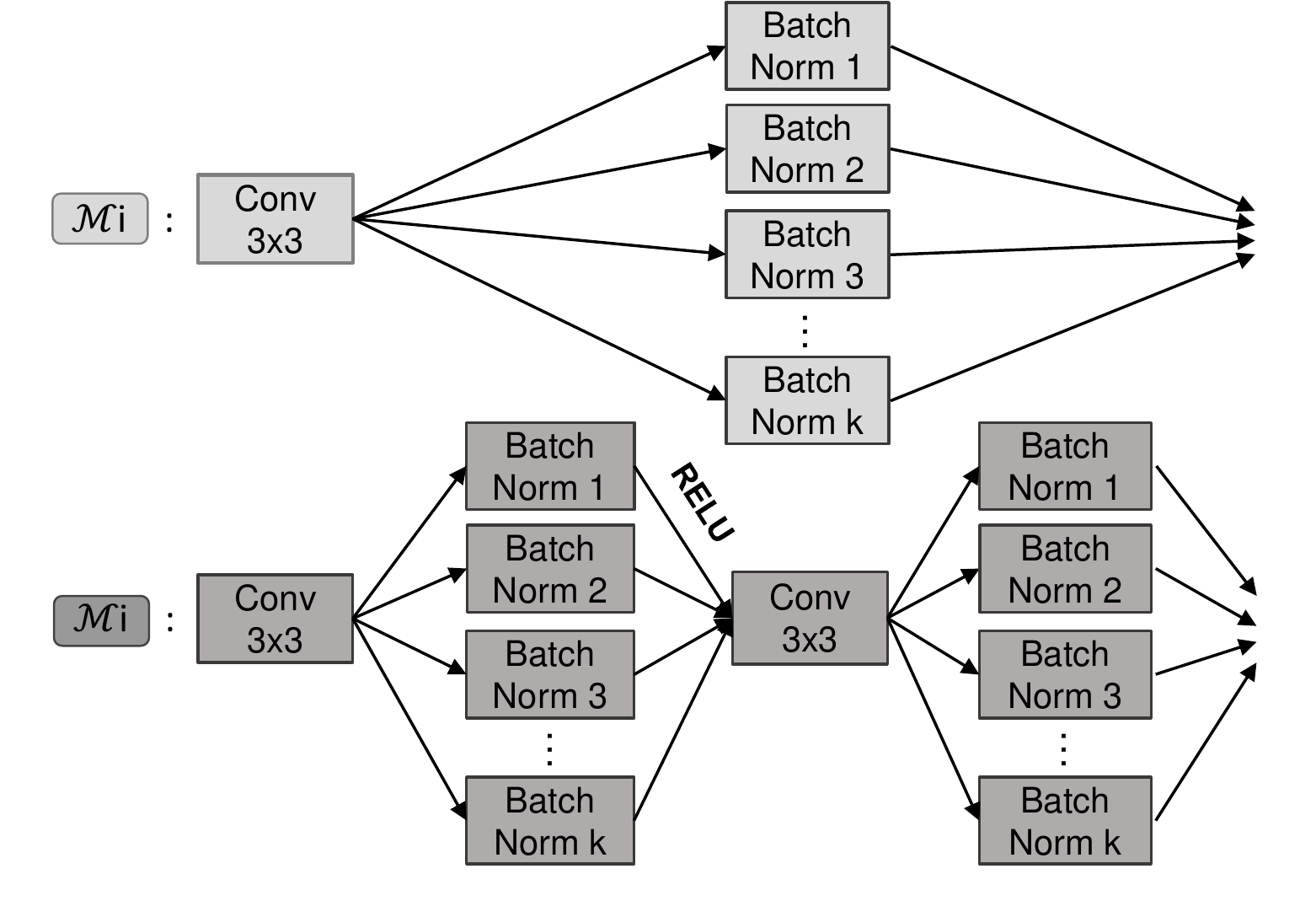}
    \caption{To handle the heterogeneity of the tasks defined in our FiveTasks experiments, each task was assigned a separate batch normalization instance in each batch normalization layer in the base network. In case a new task is introduced, a new batch normalization instance will be added to each batch normalization layer in the base network for that task.}
    \label{fig:multiplebatchnorms}
\end{figure}
%----------------------------------------------------------------------
In this experiment, we ran experiments on more heterogeneous classification tasks. We borrowed this experimental design from the work by Hung et al. \cite{hung2019compacting}. Five classification tasks were defined on five different image datasets.  Table \ref{tab:datasets} shows a summary of each dataset. To handle the heterogeneity in the input data distributions, we modified the base network by assigning a specific batch normalization module to each task in all the batch normalization layers in the base network. Fig. \ref{fig:multiplebatchnorms} shows how the batch normalization layers were modified to have multiple batch normalization modules. Once a new task is added for training, a new batch normalization module will be added to each batch normalization layer in the base network. \par
In this experiment, ResNet-50 network was used to generate the base network. A detailed representation of the base network architecture generated using ResNet-50 is shown in Appendix A of the supplementary Materials. The total number of modules in each layer was set to 8 ($M$=8). The base network was initialized with parameters of a ResNet-50 model trained on the ImageNet dataset following Hung et al. \cite{hung2019compacting}. Two modules in each layer were randomly selected and assigned to each task ($N$=2). The module sharing profile for this experiment is shown in Fig. \ref{fig:modulesharing} (Right).\par
%----------------------------------------------------------------------
\begin{table}[t]
    \centering
    \resizebox{0.8\columnwidth}{!}{%Or, if you a
    \begin{tabular}{|lcccccc|}
    \rowcolor{Gray}
    \hline
    {Method}& Task 1 & Task 2 & Task 3 & Task 4 & Task 5 &  {Avg}\\
    \hline
    & 82.75 \textpm~0.40 & 92.08 \textpm~0.26&  95.32  \textpm~0.42& 74.01 \textpm~0.88& 79.35  \textpm~0.23 & 84.70\\
     \cline{2-7}
     Single&\cellcolor{Gray} Tasks $All \backslash {1}$ & \cellcolor{Gray}Tasks $All \backslash {2}$& \cellcolor{Gray}Tasks $All \backslash {3}$& \cellcolor{Gray}Tasks $All \backslash {4}$& \cellcolor{Gray}Tasks $All \backslash {5}$& \cellcolor{Gray}\\
     \cline{2-7}
     & 0.57 \textpm~0.36& 0.60 \textpm~0.43& 0.43 \textpm~0.16& 0.57 \textpm~0.50 & 0.49 \textpm~0.18 & 0.53\\
    \rowcolor{Gray}
    \hline
    & Task 1 & Task 2 & Task 3 & Task 4 & Task 5 & \\
    \hline
    Parallel &  80.63 \textpm~0.27& 92.08 \textpm~0.29& 94.38 \textpm~0.15& 74.44 \textpm~1.02& 78.32 \textpm~0.65&  83.97\\
    \hline
    CPG \cite{hung2019compacting} & 83.59 & 92.80 & 96.62 & 77.15 & 80.33 & 86.10\\
    \hline
    \end{tabular}}
    \caption{Table compares validation accuracy of DL models trained for the FiveTasks experiment using single task learning and PaRT. For DL models trained using single task learning, the measured validation accuracy on the learned task and unlearned tasks are provided. The results are compared with those of the proposed algorithm by Hung et al.\cite{hung2019compacting} named CPG. Five different models were trained for each task for single task learning and PaRT and the validation accuracy of all the five models were averaged. }
    \label{tab:5tasks}
\end{table}
%----------------------------------------------------------------------
\begin{figure}[t] %{R}{0.6\columnwidth}
    \centering
    \begin{tabular}{c}
        \includegraphics[width=0.45\columnwidth, trim={0.85cm 0.6cm 0cm 0.5cm}, clip]{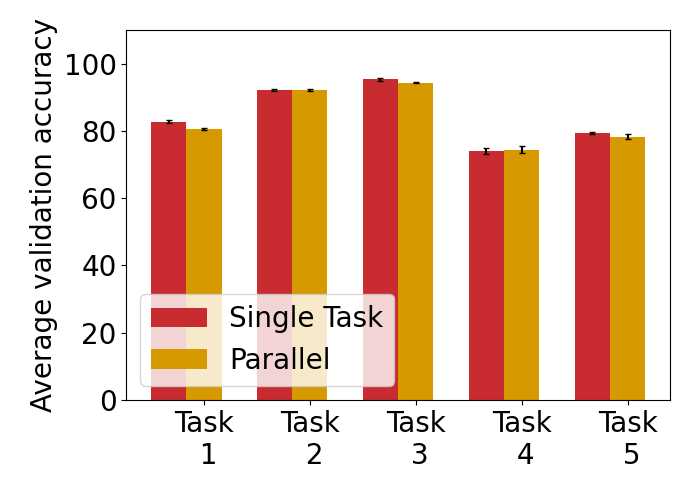}   \\
    \end{tabular}
    \caption{Validation accuracy of tasks in the FiveTasks experiments using single task learning and PaRT. The validation accuracy is measured on five DL models trained for each learning method.}
    \label{fig:5tasks}
\end{figure}
%----------------------------------------------------------------------
\begin{figure*}[!t]
\centering
\resizebox{0.98\textwidth}{!}{%
    \begin{tabular}{c}
         \includegraphics[width=\textwidth, trim={0cm 6.3cm 3cm 0cm},clip]{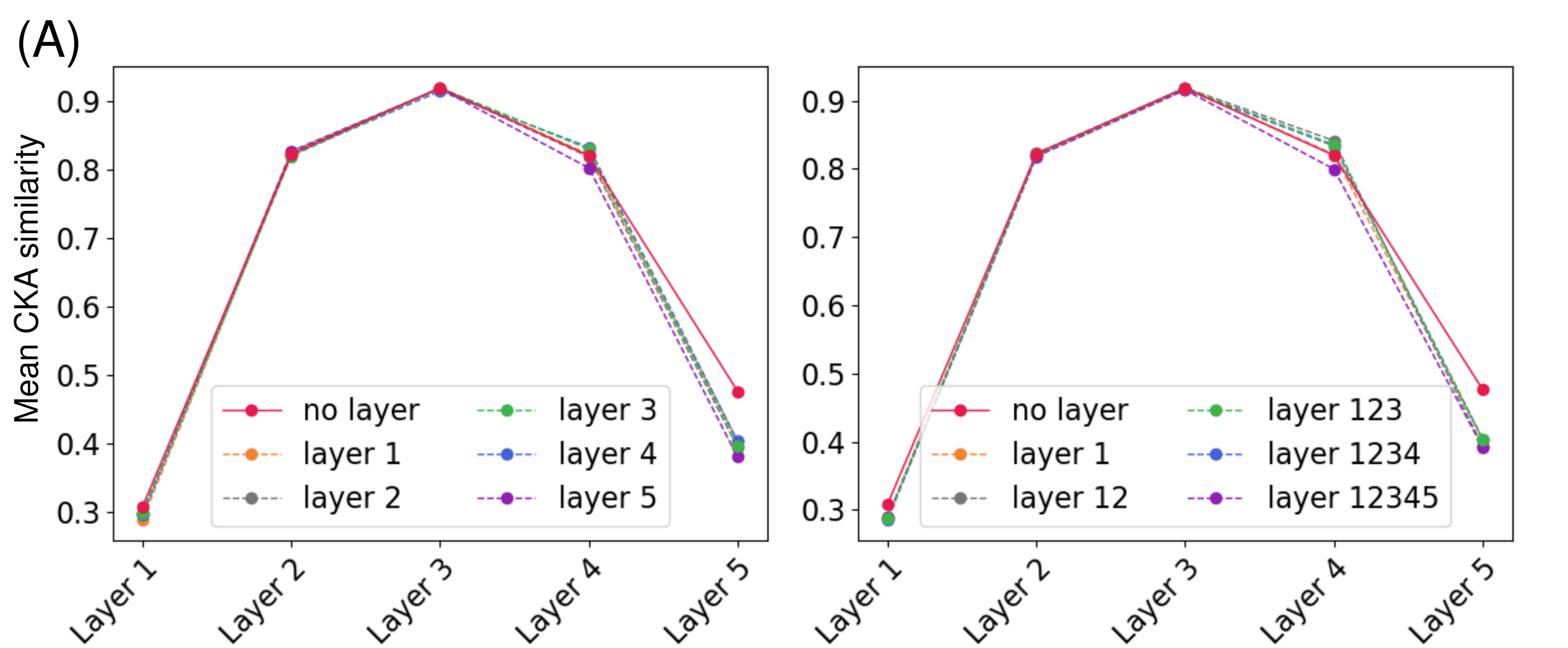} \\
         \includegraphics[width=\textwidth, trim={0cm 6.3cm 3cm 0cm},clip]{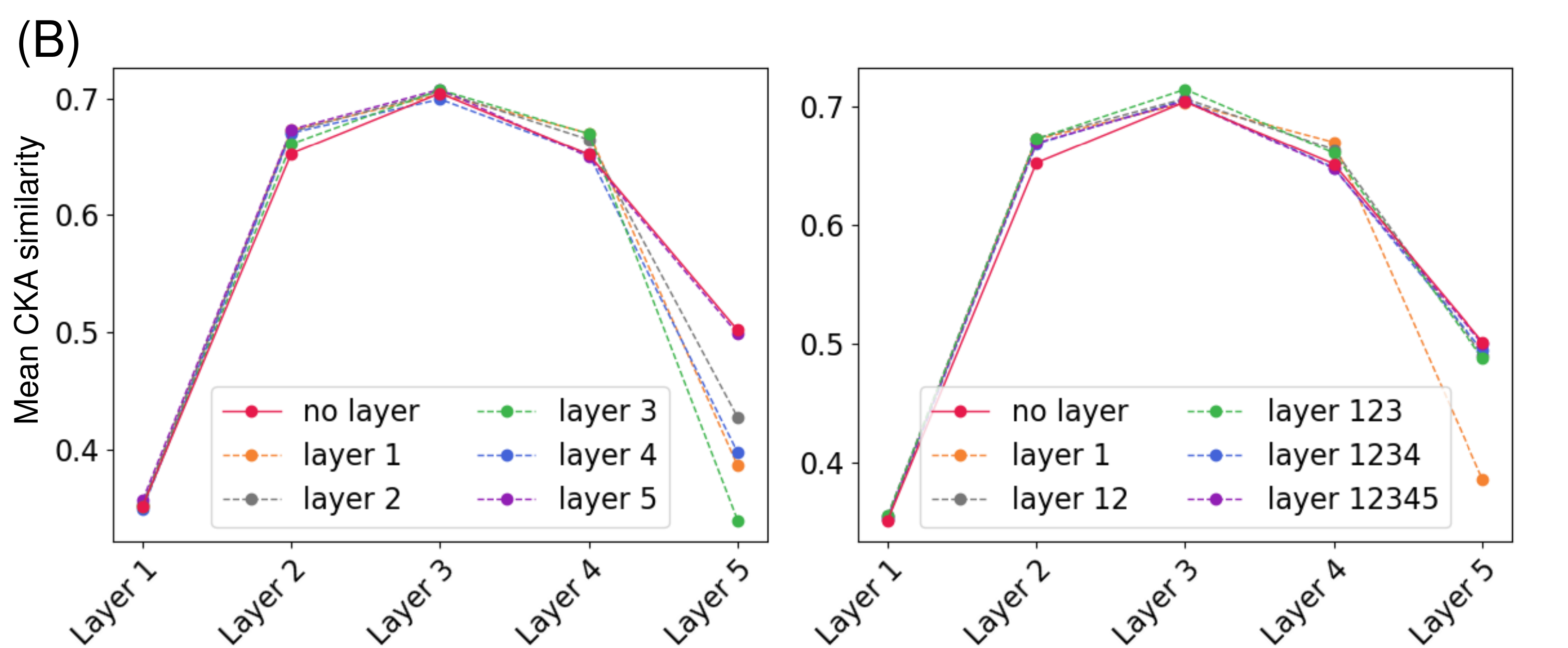} \\
    \end{tabular}
    }
    \caption{(A) Figure compares learned representations at each layer by DL models trained with PaRT on classification of the Stanford Cars and CUBs datasets with different amounts of module-sharing between the two tasks. The label 'no layer' means a DL model was trained on the two tasks while no modules were shared between the two tasks which is equivalent to single task learning. Label 'layer 1' means a DL model was trained on the two tasks while modules in layer 1 were shared between the two tasks and label 'layer 12345' means a DL model was trained on the two tasks while all modules at all layers were shared between the two tasks. (B) Figure compares learned representations by DL models trained with PaRT on classification of the Flowers and Sketches datasets with different amounts of module-sharing between the two tasks. The labels convey the same meaning as in (A).}
    \label{fig:cka_overall}
\end{figure*}
%----------------------------------------------------------------------
\begin{figure*}[t]
\centering
\resizebox{0.98\textwidth}{!}{%
    \begin{tabular}{c}
         \includegraphics[width=\textwidth, trim={0cm 9.2cm 0cm 0cm},clip]{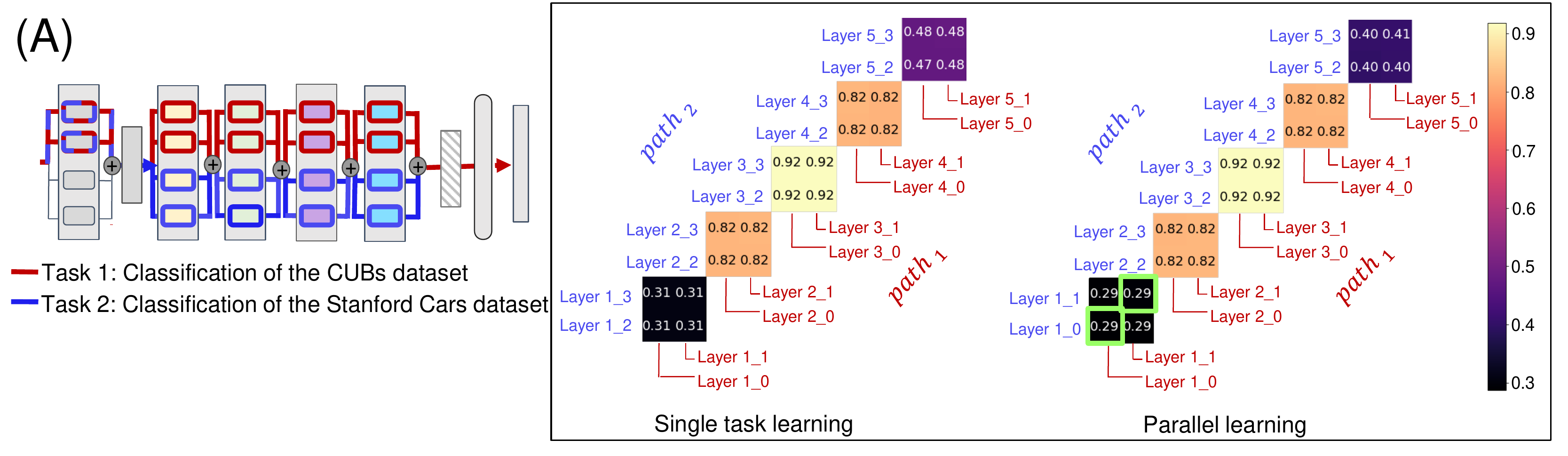} \\
         \includegraphics[width=\textwidth, trim={0cm 9.2cm 0cm 0cm},clip]{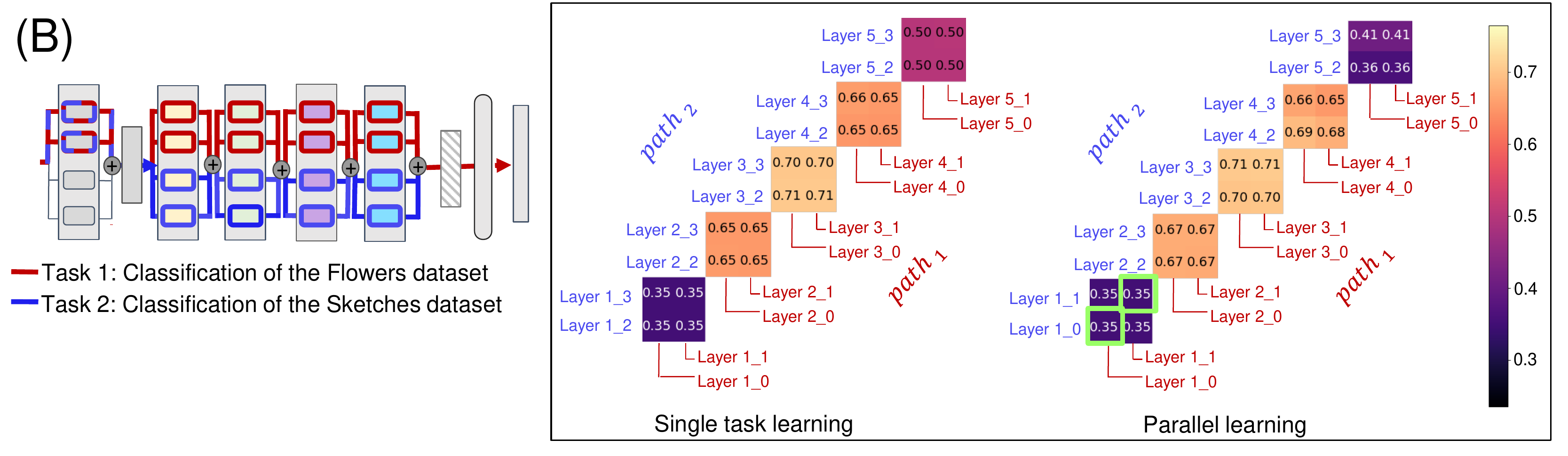} \\
    \end{tabular}
    }
    \caption{Figure compares learned representations by a DL model trained with PaRT while no modules were shared between the two tasks,which is equivalent to single task learning, with learned representations by a DL model trained with PaRT on those two tasks while sharing two modules at the first layer of the network. (A) provides this comparison for the Stanford Cars and CUBs datasets. (B) shows this comparison for the Flowers and WikiArt datasets. CKA with RBF kernel ($\sigma$=0.5) \cite{kornblith2019similarity} was used as the representation similarity metric. Higher values show more similar representations. The tasks and their selected modules are shown at the left. The shared modules are also shown by a green bounding box on the heatmaps.}
    \label{fig:cka_a}
\end{figure*}
%----------------------------------------------------------------------
\begin{figure*}[t]
\centering
\resizebox{0.98\textwidth}{!}{%
    \begin{tabular}{c}
         \includegraphics[width=\textwidth, trim={0cm 9.2cm 0cm 0cm},clip]{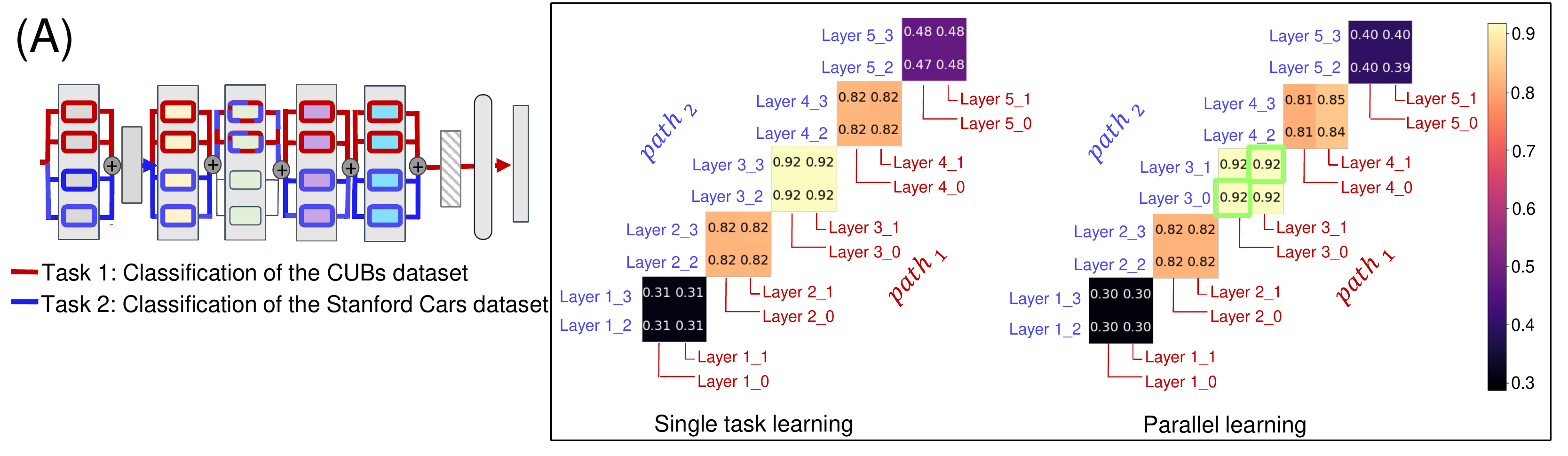} \\
         \includegraphics[width=\textwidth, trim={0cm 9.2cm 0cm 0cm},clip]{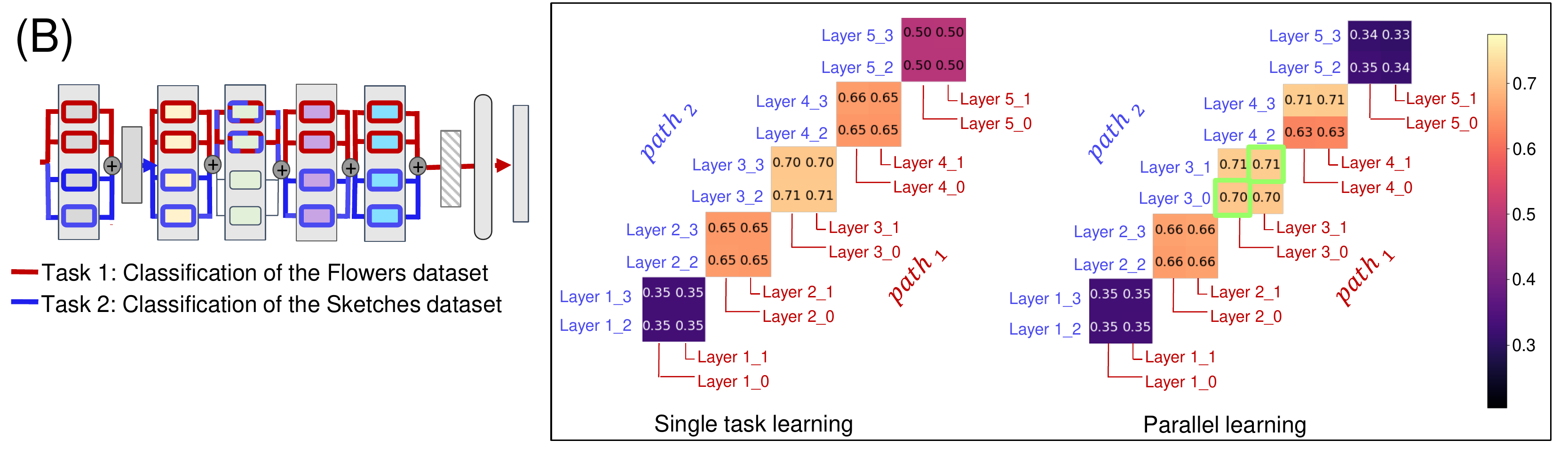} \\
    \end{tabular}
    }
    \caption{Figure compares learned representations by a DL model trained with PaRT while no modules were shared between the two tasks, which is equivalent to single task learning, with learned representations by a DL model trained with PaRT on those two tasks while sharing two modules at the third layer of the network. (A) provides this comparison for the Stanford Cars and CUBs datasets. (B) shows this comparison for the Flowers and WikiArt datasets. CKA with RBF kernel ($\sigma$=0.5) \cite{kornblith2019similarity} was used as the representation similarity metric. Higher values show more similar representations. The tasks and their selected modules are shown at the left. The shared modules are also shown by a green bounding box on the heatmaps.}
    \label{fig:cka_b}
\end{figure*}
%----------------------------------------------------------------------
\begin{figure*}[t]
\centering
\resizebox{0.98\textwidth}{!}{%
    \begin{tabular}{c}
         \includegraphics[width=\textwidth, trim={0cm 9.2cm 0cm 0cm},clip]{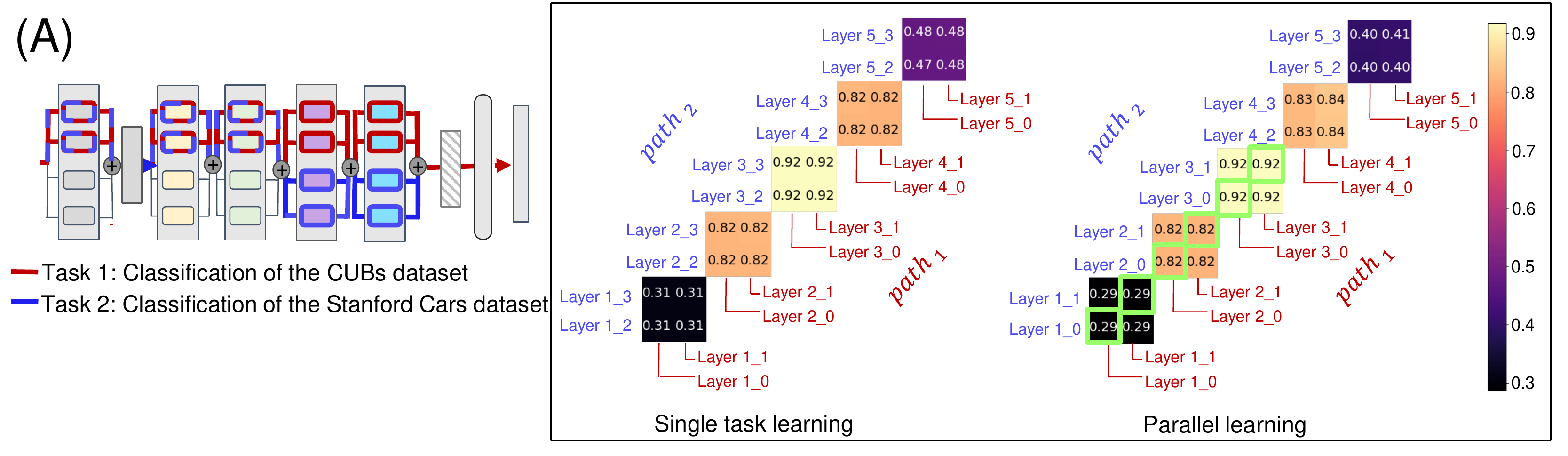} \\
         \includegraphics[width=\textwidth, trim={0cm 9.2cm 0cm 0cm},clip]{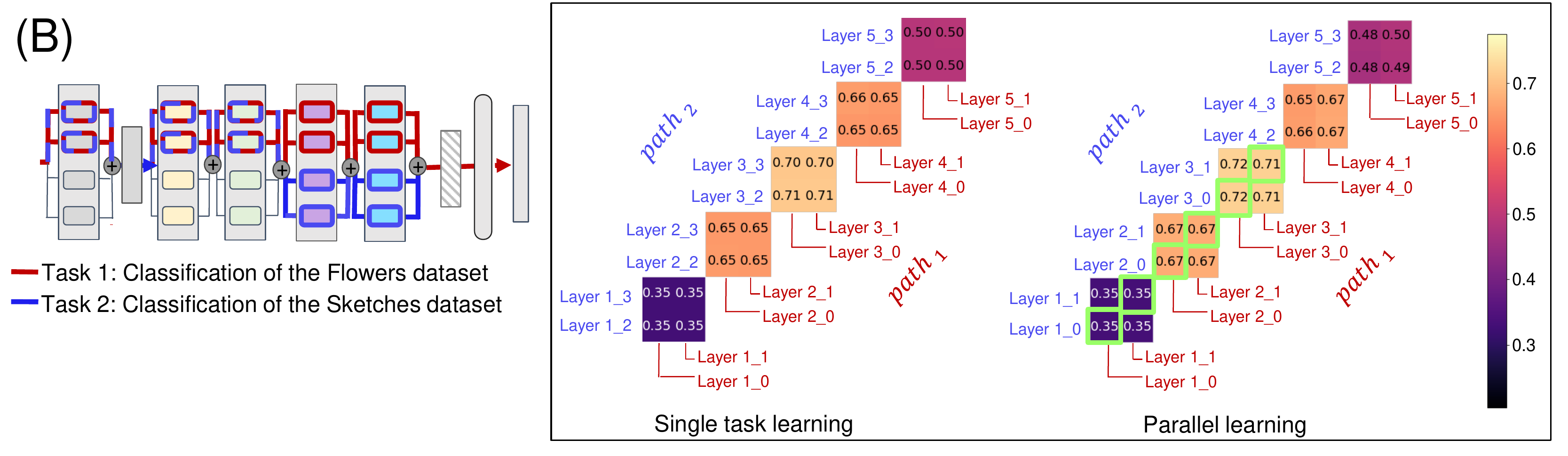} \\
    \end{tabular}
    }
    \caption{Figure compares learned representations by a DL model trained with PaRT while no modules were shared between the two tasks, which is equivalent to single task learning, with learned representations by a DL model trained with PaRT on those two tasks while sharing six modules at the first, second and third layers of the network. (A) provides this comparison for the Stanford Cars and CUBs datasets. (B) shows this comparison for the Flowers and WikiArt datasets. CKA with RBF kernel ($\sigma$=0.5) \cite{kornblith2019similarity} was used as the representation similarity metric. Higher values show more similar representations. The tasks and their selected modules are shown at the left. The shared modules are also shown by a green bounding box on the heatmaps.}
    \label{fig:cka_c}
\end{figure*}
%----------------------------------------------------------------------
\begin{figure*}[t]
\centering
\resizebox{0.98\textwidth}{!}{%
    \begin{tabular}{c}
         \includegraphics[width=\textwidth, trim={0cm 9.2cm 0cm 0cm},clip]{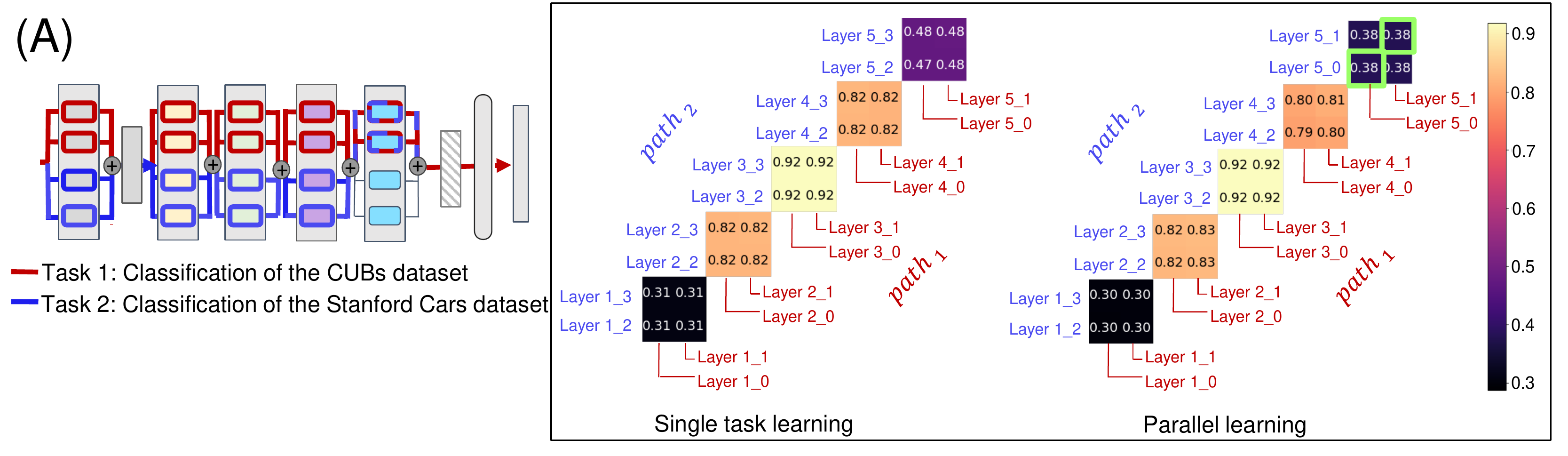} \\
         \includegraphics[width=\textwidth, trim={0cm 9.2cm 0cm 0cm},clip]{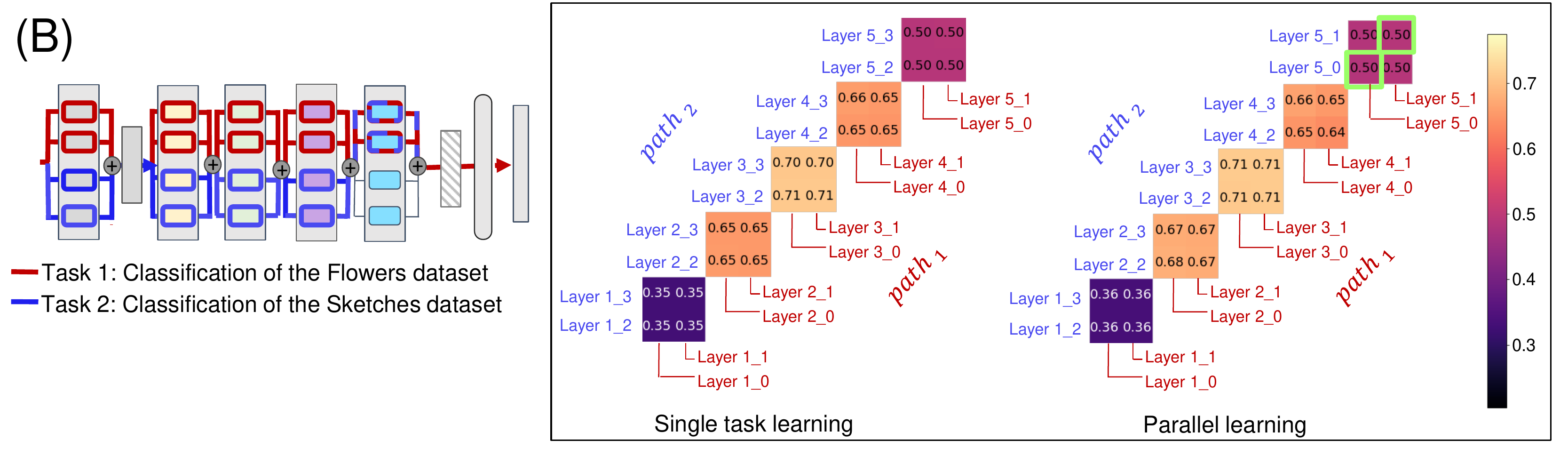} \\
    \end{tabular}
    }
    \caption{Figure compares learned representations by a DL model trained with PaRT while no modules were shared between the two tasks, which is equivalent to single task learning, with learned representations by a DL model trained with PaRT on those two tasks while sharing two modules at the fifth layer of the network. (A) provides this comparison for the Stanford Cars and CUBs datasets. (B) shows this comparison for the Flowers and WikiArt datasets. CKA with RBF kernel ($\sigma$=0.5) \cite{kornblith2019similarity} was used as the representation similarity metric. Higher values show more similar representations. The tasks and their selected modules are shown at the left. The shared modules are also shown by a green bounding box on the heatmaps.}
    \label{fig:cka_d}
\end{figure*}
%----------------------------------------------------------------------
Since PaRT outperformed sequential learning in the previous experiments, we compared PaRT solely with single task learning in this section. Fig. \ref{fig:5tasks} shows the average validation accuracy for each task over five DL models trained with different seeds using single task learning and PaRT. It is shown that the average validation accuracy of DL models trained using PaRT is only slightly lower than that of DL models trained with single task learning. Table \ref{tab:5tasks} shows the mean and standard deviation of validation accuracy for each task for models trained using single task learning and PaRT. For Task 4, the mean validation accuracy of PaRT is even higher than that of DL models trained using single task learning. We have also shown the validation accuracy achieved by Hung et al. \cite{hung2019compacting}. CPG \cite{hung2019compacting} which is a more complex CL algorithm outperforms PaRT with an average of 2.13\% in validation accuracy. Considering the simplicity of our  base network and the current strategy for module assignment, we believe a better base network and a more efficient module assignment strategy may train models that outperform the performance of models trained using CPG \cite{hung2019compacting}.
\section{Discussion}
In sections \ref{sec:homo_cifar100} and \ref{sec:hetero_cifar10_cifar100}, we showed that DL models trained using PaRT can utilize network resources more efficiently compared to models trained with the sequential learning algorithm. We also showed that DL models trained with PaRT on more heterogeneous tasks in section \ref{sec:hetero_5tasks} can perform as well as models trained with single task learning. In order to explain how PaRT enables these improvements in network resource efficiency and task performance compared to sequential learning, we compared the task-related and shared representations learned by the DL modules for each task. 
\subsection{Learned Representations}
We designed a separate experiment to study the effect of module-sharing on the learned representations by the base network using PaRT. For this experiment, we considered a base network using the ResNet-50 architecture with four modules in each layer ($M$=4). Similar to the FiveTasks experiment, we initialized the network with parameters of a ResNet-50 model trained on the ImageNet dataset. The base network was trained on two tasks only. Two modules in each layer were selected and assigned to each task ($N$=2), however, selection of the modules was not done randomly. Ten different setups were considered with each setup defining a specific way of sharing modules between the two tasks. For instance, in 'no layer' setup, no modules were shared between the two tasks. This setup is equivalent to single task learning of each of the two tasks. In 'layer 1' setup, two modules in layer 1 were shared between the two tasks. and in 'layer 12345' setup, two modules at each layer were shared between the two tasks. In experiment (A), the base network was trained using PaRT on two tasks which were classification of Stanford Cars and CUBs datasets. In experiment (B), the base network was trained using PaRT on two tasks which were classification of Flowers and Sketches datasets. Three DL models with different random seeds were trained using each setup  and the similarity of their learned representations for the two tasks were compared with each other in both the experiments. \par 
We deployed the representation similarity metric proposed by Kornblith et al. \cite{kornblith2019similarity} called CKA. For this comparison, 1000 samples from the validation dataset of each task with equal number of samples per class were passed through each DL model and the similarity of the representations for the two tasks at each layer were compared with DL models trained with ten different setups. Fig. \ref{fig:cka_overall} shows the average similarity of learned representations by the modules for the two tasks  at each layer for different module-sharing setups for experiments (A) and (B). Multiple important points can be observed in these plots including: 
\begin{itemize}
    \item Learned representations at the middle layers are more similar across the two tasks compared to learned representations at the first and last layers. 
    \item Compared to single task learning, sharing modules at layer 3 results in more similar representations across the two tasks at the surrounding layers but not at the first and last layers. 
    \item Compared to single task learning, sharing modules at layers 1, 2, and 3 results in more similar representations across the two task at layer 4 but not at the last layer.
    \item  Compared to other module-sharing setups, sharing modules at all layers leads to less similar representations at different layers across the two tasks.
\end{itemize}

We have also shown the selected path for each task and the average pair-wise CKA similarity between different modules at the same layer for DL models trained using 'no layer', 'layer 1', 'layer 3', 'layer 123', and 'layer 5' in Figures \ref{fig:cka_a}, \ref{fig:cka_b}, \ref{fig:cka_c} and \ref{fig:cka_d}, respectively. The average CKA similarity metric is measured across 3 DL models trained using each setup with different random seeds. We show the CKA comparison results using an RBF kernel with $\sigma$=0.5 bandwidth which controls the extent to which similarity of small distances is emphasized over large distances. A value close to one, shown by a bight yellow color, means more similar representations were learned for the two tasks by the corresponding module. The shared modules between the two tasks in DL models trained using PaRT are highlighted with a green bounding box. This experiment provides a high level of transparency over what features a DL model learns at different layers. We believe the low CKA similarity seen across different tasks at the first and final layers implies that the basic features learned by DL models for different tasks and the final decision making is quite different across tasks whereas the abstract features extracted in the middle layers of the network are more similar across different tasks. 
\vspace{-0.5em}
\subsection{Limitations}
In a normal training scenario, a single ResNet-18 or ResNet-50 model usually suffices to learn one task. However, PaRT currently assigns $N > 1$ modules in each layer to a task. In an ideal scenario, PaRT should be able to learn multiple tasks using almost the same amount of network resources that would be required while training a DL model on each task. Reducing utilization of network resources is a part of our future plan for extending PaRT. PaRT should also be tested on more diverse tasks such as classification and segmentation tasks. Moreover, while the random selection of modules in the paths for different tasks already gives decent results, this process of selecting modules can be optimized by incorporating Reinforcement Learning which takes the average accuracy or loss as the reward for the agent to learn.  
\section{Conclusion}
In this paper, we take a parallel learning approach to train DL models on multiple tasks simultaneously. Parallel learning achieves a better performance than many sequential continual learning methods and state-of-the-art methods that are proposed in other domains. This is the first effort that harnesses shared representations to improve individual task performance and increases efficient use of network resources. It also increases transparency in the model, therefore allowing us to draw explanations out of the model's predictions.  Parallel learning has widespread applications in the real-world. One area where it can be harnessed is the healthcare industry. Our model enables a single DL model to have the capability of identifying multiple diseases such as lung cancer, and heart failure, hence largely lowering the cost of incorporating DL in hospitals. 

{\small
\bibliographystyle{ieee}
}

\section{APPENDIX A: Our Base Network Generated Using The ResNet-50 Network Architecture}
\begin{figure}[h]
    \centering
    \includegraphics[width=\textwidth, trim={0cm 0cm 5.5cm 0cm},clip]{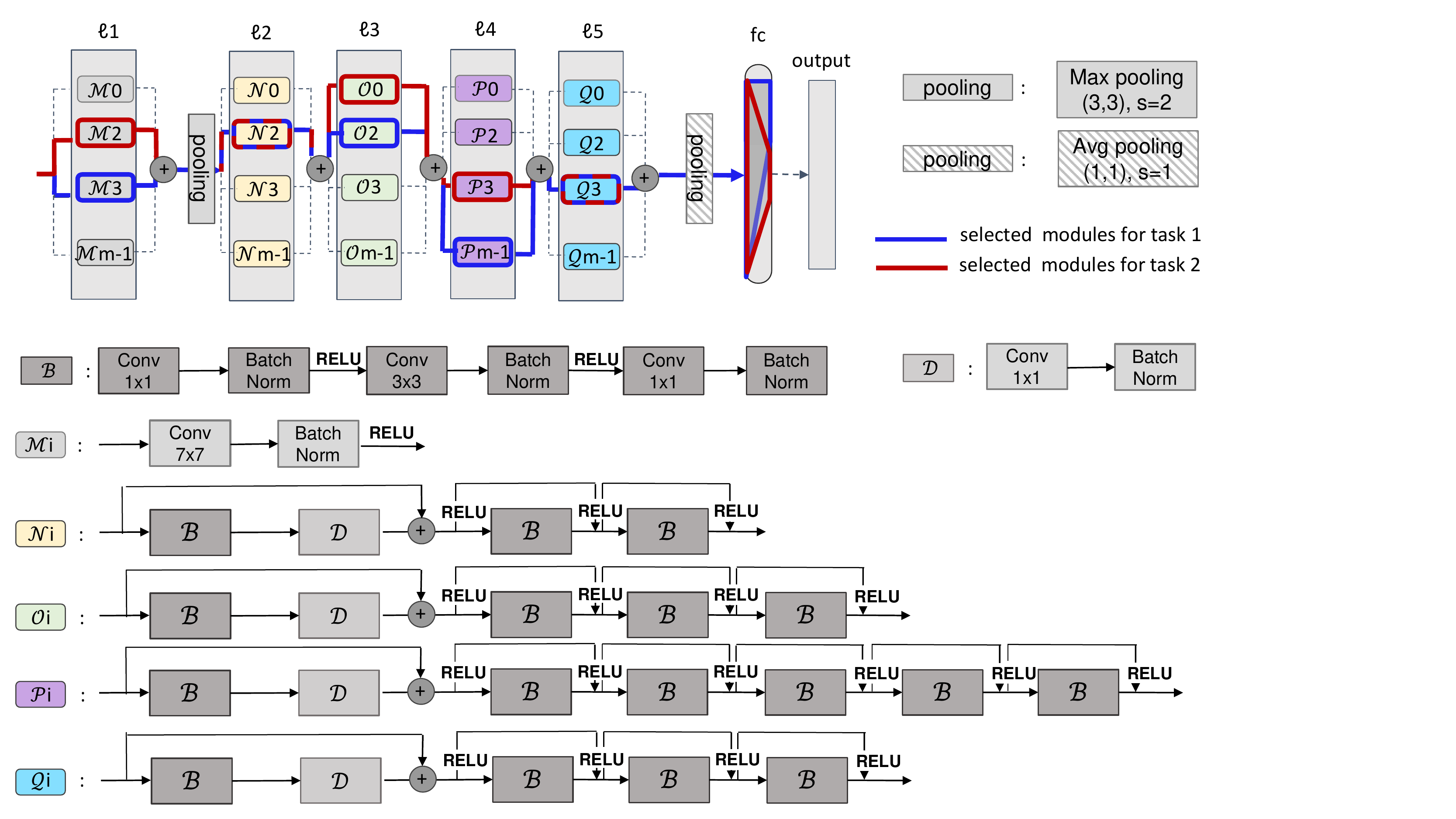}
    \caption{Figure shows the architecture of the base network that was deployed for the FiveTasks experiments in section 4.2.2 in the original paper. One row in the base network is equivalent to the ResNet-50 network architecture.}
    \label{fig:base_net2}
\end{figure}

Fig \ref{fig:base_net2} shows the architecture of the base network generated using the ResNet-50 network architecture. This base network architecture was deployed in the FiveTasks experiments presented in section 4.2.2 in the original paper. Similar to the base network architecture generated using ResNet-18, the base network consists of multiple layers, with a total of $M$ modules in each layer. The outputs of modules in the same layer are summed together, and the sum is fed to all the modules in the next layer. One row in the base network is equivalent to the ResNet-50 network architecture. The architecture of each module is shown at the bottom of the figure. 

\section{APPENDIX B: Experiment Setups and Hyperparameter Values}
In this section, we provide the experiment setup for each learning method for all three experiments conducted in the original paper and the hyperparameters we eventually decided on after fine-tuning them. 
%--------------------------------------------------------------------------
\subsection{CIFAR100 Experiments}
This section provides the experiment setup for each compared learning method in section 4.1 in the original paper. 
\label{sec:cifar100}
\subsubsection{Image Pre-processing Pipeline}
The same image pre-processing pipeline was applied to the images before running the learning methods mentioned below. We normalized images in the CIFAR100 training dataset using mean values of $\{0.507, 0.486, 0.44\}$ and standard deviation values of $\{0.267, 0.256, 0.276\}$ for the $RGB$ channels, respectively. We also applied image augmentation methods including $RandomPerspective$ with probability of 0.75 and distortion scale of 0.2, $RandomHorizontalFlip$ and $RandomCrop$ with output size of 32 pixels and 4 pixels padding. The preprocessing pipeline for the images in the CIFAR100 test dataset included only image normalization using the same mean and standard deviation values.
\subsubsection{Single Task Learning}
In this experiment, we used the base network generated using the ResNet-18 network architecture. At each layer, $M$=12 modules were specified. For each task, $N$=4 modules were randomly selected and were assigned to the task. $ADAM$ optimizer with $Cross Entropy$ loss function was deployed. The initial learning rate for the optimizer was set to 0.001. The learning rate was halved at epochs 20, 30, and 40. Models were trained until convergence.
\subsubsection{Sequential Learning}
For this experiment, the same hyperparameter values and experiment setups were followed as for the single task learning experiments. During each run, the classes assigned to each task and the order of tasks to learn were set randomly. The model was trained on each task sequentially. The selected modules for each task were frozen after the training for the task ended.
\subsubsection{Parallel Learning}
The hyperparameter values were set to the same values used in the single task learning experiments and the $batch\text{-}set$ size was set to 10.
\subsubsection{MRN \cite{long2015learning}}
We used the official code provided by the authors. The original code uses VGG-16 as the underlying network architecture, and the last fully connected layer in VGG-16 is used as the head for each task. We replaced this network with ResNet-18, and the last fully connected layer in ResNet-18 as the head for each task. The initial learning rate was set to 0.05, the gamma value was set to 0.2 with 0.75 power and 3000 stepsize.
\subsubsection{pDarts \cite{chen2019progressive}}
We used the official code provided by the authors. The initial learning rate is set to 0.025 with momentum of 0.9 and weight decay of $3\times10^{-4}$. The number of search epochs is set to 5 and training epochs to 60. We specified 16 initial search channels and 36 initial train channels. The total number of search layers was set to 5, and the total number of train layers was set to 20. We did not use cutout. Drop path probability and training portion were set to 0.3. The initial $DropOut$ probability was set to 0.0, 0.3, 0.6 for stage 1, 2, and 3, respectively. An $ADAM$ optimizer with a learning rate of 0.0006 and weight decay of 0.001 and momentum of (0.5,0.999) were used for architecture encoding. Finally, gradient clipping was set to 5.
\subsubsection{ENAS \cite{pham2018efficient}}
For ENAS experiments, we used the Pytorch implementation of this algorithm on \href{https://github.com/MengTianjian/enas-pytorch}{GitHub: ENAS-Pytorch}. The parameters were trained with a momentum value of 0.9 where the learning rate followed the cosine schedule with $l_{max}=0.05$, $l_{min}=0.0005$, $T_0=10$, $T_{mul}=2$, and a weight decay of $10^{-4}$. The policy parameters were trained using $ADAM$ optimizer with a learning rate of 0.0035, where a TanH constant of 1.10, TanH reduce rate of 2.5, and a temperature of 5.0 were applied to the controller's logits. The controller entropy was added to the reward, weighted by 0.0001. The LSTM size was set to 64, with 1 layer, 0 keep probability.
\subsubsection{MANAS \cite{carlucci2019manas}}
We received the MANAS code from the authors. For MANAS experiments we set the number of layers to 20, reward baseline to 5, number of iterations to 390, the starting temperature to 1000, the ending temperature to 200, and the number of initial channels to 16. 
\subsubsection{EWC \cite{kirkpatrick2017overcoming}}
We used the implementation of EWC \cite{kirkpatrick2017overcoming} that is applied to the CIFAR100 dataset by Chaudhry et al. \cite{chaudhry2018riemannian}. We trained the ResNet-18 network architecture with multiple heads similar to our approach and used the same hyperparameters specified in the paper for the CIFAR100 dataset  that is optimizer=$ADAM$, lambda=1000, arch=$ResNet\text{-}B$ and method=$RWalk$.
\subsubsection{RPS \cite{rajasegaran2019random}}
For RPS experiments, we used our base network generated using ResNet-18 network architecture. It is important to mention that our base network is different from the network deployed in this paper. In this paper, the authors have added one extra module at layers 4, 6, and 8 in the network. These extra modules are never frozen and the output of the modules for all tasks pass through these extra modules. We set he number of modules at each layer to $M$=12 and the number of randomly selected modules for each task to $N$=4. We used the same hyperparameter values used in this paper, which include an $ADAM$ optimizer with $t_e = 2$ and an initial learning rate of $10^{-3}$ that is halved at epochs 20, 40 and 60. The scaling factor and the path selection frequency ($J$) values for the controller are set to 2.5 and 1, respectively.

%-------------------------------------------------------------------
\subsection{CIFAR10 and CIFAR100 Experiments}
This section provides the experimental setup for each compared learning method in section 4.2.1 in the original paper. 
\subsubsection{Image Pre-processing Pipeline}
The same image pre-processing pipeline was applied to the images before running the learning methods mentioned below. The same processing steps mentioned in section \ref{sec:cifar100} for the CIFAR100 dataset were deployed. For the CIFAR10 dataset, we normalized images using mean values of $\{0.4914, 0.4822, 0.4465\}$ and standard deviation values of $\{0.2023, 0.1994, 0.2010\}$ for the $RGB$ channels, respectively. The image augmentation methods used for the CIFAR100 training dataset were applied to the CIFAR10 training dataset.
\subsubsection{Single Task Learning}
In this experiment, we used the base network generated using the ResNet-18 network architecture. At each layer, $M$=15 modules were specified. For each task, $N$=4 modules were randomly selected and were assigned to the task. $ADAM$ optimizer with $Cross Entropy$ loss function was deployed. The initial learning rate for the optimizer was set to 0.001. The learning rate was halved at epochs 20, 30, and 40. Models were trained until convergence.
\subsubsection{Sequential Learning}
For this experiment, the same hyperparameter values and experiment setups were followed as for the single task learning experiments. During each run, the classes assigned to each task and the order of tasks to learn were set randomly. The model was trained on each task sequentially. The selected modules for each task were frozen after training for the task ended.
\subsubsection{Parallel Learning}
The hyperparameter values were set to the same values used in the single task learning experiments and the $batch\text{-}set$ size was set to 10.
\subsubsection{MRN \cite{long2015learning}} 
We used the same hyperparameter values that were specified for the CIFAR100 experiments.
\subsubsection{pDarts \cite{chen2019progressive}}
We used the same hyperparameter values that were specified for the CIFAR100 experiments.
\subsubsection{ENAS \cite{pham2018efficient}}
We used the same hyperparameter values that were specified for the CIFAR100 experiments.
\subsubsection{MANAS \cite{carlucci2019manas}}
We used the same hyperparameter values that were specified for the CIFAR100 experiments.
\subsubsection{EWC \cite{kirkpatrick2017overcoming}}
We used the same hyperparameter values that were specified for the CIFAR100 experiments.
\subsubsection{RPS \cite{rajasegaran2019random}}
For this experiment, we set the number of modules at each layer to $M$=15. The rest of hyperparameter values are the same as the values specified for the CIFAR100 experiments.
%--------------------------------------------------------------------------
\subsection{FiveTasks Experiments}
This section provides the experimental setup for each compared learning method in section 4.2.2 in the original paper. 
\subsubsection{Image Pre-processing Pipeline}
The same image pre-processing pipeline was applied to the images before running the learning methods mentioned below. We applied the same image pre-processing steps described by Hung et al. \cite{hung2019compacting}. The processing steps are as follows: For all five image datasets, the images in the training dataset were normalized using mean values of $\{0.485, 0.456, 0.406\}$ and standard deviation values of $\{0.229, 0.224, 0.225\}$. Image augmentation methods including $RandomHorizontalFlip$, $RandomResizedCrop$ with crop size of 224 were applied on images in the training dataset. Images in the validation dataset were normalized using the same mean and standard deviation and were resized to image size of 224 by 224 pixels. \par

\subsubsection{Single Task Learning}
In this experiment, we used the base network generated using the ResNet-50 network architecture. At each layer, $M$=8 modules were specified. For each task, $N$=2 modules were randomly selected and were assigned to the task. The modules were initialised with a ResNet-50 model trained on the ImageNet dataset. $ADAM$ optimizer with $Cross Entropy$ loss function was deployed. The initial learning rate for the optimizer was set to 0.0001. The learning rate was halved at epochs 20, 30, and 40. Models were trained until convergence. \par

\subsubsection{Parallel Learning}
The hyperparameter values were set to the same values used in the single task learning experiments and the $batch\text{-}set$ size was set to 50.\par\par

\section{APPENDIX C: Validation Accuracy for All the Compared Algorithms}
We provided the mean and standard deviation of validation accuracy for DL models trained using single task learning, sequential learning and PaRT for CIFAR100 experiments in sections 4.1 and CIFAR10 and CIFAR100 experiments in section 4.2 in table 1 and table 2 in the original paper. In this section, we provide the mean and standard deviation of validation accuracy for DL models trained with algorithms that PaRT was compared to, namely, MRN \cite{long2015learning},  pDarts \cite{chen2019progressive}, ENAS \cite{pham2018efficient}, MANAS \cite{carlucci2019manas},  EWC \cite{kirkpatrick2017overcoming} and RPS \cite{rajasegaran2019random}. The mean and standard deviation of validation accuracy were measured for five DL models trained using each learning method. 
\subsection{CIFAR100 Experiments}
Table \ref{tab:cifar100} shows the mean and standard deviation of DL models trained with RN \cite{long2015learning},  pDarts \cite{chen2019progressive}, ENAS \cite{pham2018efficient}, MANAS \cite{carlucci2019manas},  EWC \cite{kirkpatrick2017overcoming} and RPS \cite{rajasegaran2019random} for the tasks defined on the CIFAR100 dataset in section 4.1 in the original paper. For each algorithm, five DL models were trained, and the mean and standard deviation of their validation accuracy were measured. 
\begin{table}[H]
    \centering
    \begin{tabular}{|lccccccccccc|}
    \rowcolor{Gray}
    \hline
    {Learning Method} & Task 1 & Task 2 & Task 3 & Task 4 & Task 5 & Task 6 & Task 7 & Task 8 & Task 9 & Task 10 & {Avg}\\
    \hline
     \multirow{2}{*}{MRN \cite{long2015learning}} & 39.64& 38.06& 40.12& 35.24& 40.58& 38.98& 37.58& 29.76& 38.50& 37.60& \multirow{2}{*}{38.53}\\
     & \footnotesize{\textpm~7.04} & \footnotesize{\textpm~4.98}& \footnotesize{\textpm~5.56}& \footnotesize{\textpm~3.18}& \footnotesize{\textpm~3.51}& \footnotesize{\textpm~9.17}& \footnotesize{\textpm~4.07}& \footnotesize{\textpm~2.33}& \footnotesize{\textpm~7.05}& \footnotesize{\textpm~8.22}& \\
     \hline
     \multirow{2}{*}{pDarts \cite{chen2019progressive}} & 75.56& 70.20& 73.64& 64.98& 76.96& 77.32& 75.80& 72.88& 63.44& 79.92& \multirow{2}{*}{71.10}\\
     & \footnotesize{\textpm~6.36} & \footnotesize{\textpm~6.59}& \footnotesize{\textpm~13.84}& \footnotesize{\textpm~16.87}& \footnotesize{\textpm~7.77}& \footnotesize{\textpm~2.64}& \footnotesize{\textpm~6.29}& \footnotesize{\textpm~6.49}& \footnotesize{\textpm~10.77}& \footnotesize{\textpm~2.94}& \\
    \hline
    \multirow{2}{*}{ENAS \cite{pham2018efficient}} & 96.00& 84.00& 92.00& 94.00& 86.00& 94.00& 94.00& 90.00& 90.00& 70.00& \multirow{2}{*}{87.00}\\
      & \footnotesize{\textpm~4.90} & \footnotesize{\textpm~4.90}& \footnotesize{\textpm~11.66}& \footnotesize{\textpm~4.90}& \footnotesize{\textpm~4.90}& \footnotesize{\textpm~4.90}& \footnotesize{\textpm~8.00}& \footnotesize{\textpm~6.32}& \footnotesize{\textpm~0.00}& \footnotesize{\textpm~6.32}& \\
    \hline
    \multirow{2}{*}{MANAS \cite{carlucci2019manas}} & 88.96& 84.38& 88.32& 86.84& 89.02& 87.68& 87.18& 84.64& 85.96& 88.58& \multirow{2}{*}{85.94}\\
      & \footnotesize{\textpm~2.60} & \footnotesize{\textpm~3.65}& \footnotesize{\textpm~2.18}& \footnotesize{\textpm~1.50}& \footnotesize{\textpm~2.35}& \footnotesize{\textpm~3.38}& \footnotesize{\textpm~1.38}& \footnotesize{\textpm~3.31}& \footnotesize{\textpm~6.21}& \footnotesize{\textpm~2.83}& \\
    \hline
    \multirow{2}{*}{EWC \cite{kirkpatrick2017overcoming}} & 69.54& 52.14& 51.78& 50.36	& 55.78& 51.24&53.88 & 46.72& 51.70& 53.52& \multirow{2}{*}{53.67} \\
      & \footnotesize{\textpm~2.39} & \footnotesize{\textpm~2.56}& \footnotesize{\textpm~4.90}& \footnotesize{\textpm~2.25}& \footnotesize{\textpm~3.37}& \footnotesize{\textpm~4.64}& \footnotesize{\textpm~3.45}& \footnotesize{\textpm~4.03}& \footnotesize{\textpm~4.05}& \footnotesize{\textpm~4.37}& \\
    \hline
    \multirow{2}{*}{RPS \cite{rajasegaran2019random}}& 39.02& 30.18& 60.12& 68.84& 61.30& 66.24& 65.18& 70.54& 66.46& 68.30 &  \multirow{2}{*}{62.11}\\
      & \footnotesize{\textpm~18.67} & \footnotesize{\textpm~13.30}& \footnotesize{\textpm~7.92}& \footnotesize{\textpm~8.80}& \footnotesize{\textpm~8.26}& \footnotesize{\textpm~7.30}& \footnotesize{\textpm~12.46}& \footnotesize{\textpm~5.45}& \footnotesize{\textpm~6.12}& \footnotesize{\textpm~3.70}& \\
    \hline
    \end{tabular}
    \caption{Table compares the validation accuracy of DL models trained with different learning methods on 10 tasks defined on the CIFAR100 dataset. Definition of the tasks are provided in section 4.1 in the original paper. For each learning method, five DL models were trained, and the mean and standard deviation of validation accuracy for all five models are shown.}
    \label{tab:cifar100}
\end{table}

\subsection{CIFAR10 and CIFAR100 Experiments}
Table \ref{tab:cifar100} shows the mean and standard deviation of DL models trained with RN \cite{long2015learning},  pDarts \cite{chen2019progressive}, ENAS \cite{pham2018efficient}, MANAS \cite{carlucci2019manas},  EWC \cite{kirkpatrick2017overcoming} and RPS \cite{rajasegaran2019random} for the tasks defined on the CIFAR10 and CIFAR100 datasets in section 4.2 in the original paper. For each algorithms, five DL models were trained, and the mean and standard deviation of their validation accuracy were measured.
\begin{table}[H]
    \centering
    \begin{tabular}{|lccccccccccc|}
    \rowcolor{Gray}
    \hline
    {Learning Method} & Task 1 & Task 2 & Task 3 & Task 4 & Task 5 & Task 6 & Task 7 & Task 8 & Task 9 & Task 10 & {Avg}\\
    \hline
     \multirow{2}{*}{MRN \cite{long2015learning}} & 34.37 & 57.20	& 50.85	& 63.56	& 70.40	& 52.02	& 70.05	& 51.73	& 50.34	& 65.94	& 59.00	 \\
     & \footnotesize{\textpm~3.07} & \footnotesize{\textpm~18.40} & \footnotesize{\textpm~18.66} & \footnotesize{\textpm~25.73} & \footnotesize{\textpm~18.98} & \footnotesize{\textpm~26.99} & \footnotesize{\textpm~17.89} & \footnotesize{\textpm~19.28} & \footnotesize{\textpm~21.37} & \footnotesize{\textpm~27.39} & \footnotesize{\textpm~25.92} \\
     \hline
     \multirow{2}{*}{pDarts \cite{chen2019progressive}} & 70.80	& 85.36	& 83.99	& 88.08	& 93.12	& 82.30	& 91.53	& 83.22	& 77.45	& 87.51	& 83.77\\
     & \footnotesize{\textpm~7.46} & \footnotesize{\textpm~13.53} & \footnotesize{\textpm~10.03} & \footnotesize{\textpm~11.38} & \footnotesize{\textpm~9.19} & \footnotesize{\textpm~11.95} & \footnotesize{\textpm~11.44} & \footnotesize{\textpm~11.44} & \footnotesize{\textpm~14.34} & \footnotesize{\textpm~13.82} & \footnotesize{\textpm~15.60} \\
    \hline
    \multirow{2}{*}{ENAS \cite{pham2018efficient}} & 68.80	& 66.40	& 58.00	& 64.80	& 73.20	& 71.60	& 77.60	& 64.00	& 62.00	& 66.40	&  70.00\\
     & \footnotesize{\textpm~10.48} & \footnotesize{\textpm~9.58} & \footnotesize{\textpm~16.00} & \footnotesize{\textpm~17.09} & \footnotesize{\textpm~19.00} & \footnotesize{\textpm~20.06} & \footnotesize{\textpm~20.18} & \footnotesize{\textpm~14.97} & \footnotesize{\textpm~13.27} & \footnotesize{\textpm~12.80} &  \footnotesize{\textpm~15.49} \\
    \hline
    \multirow{2}{*}{MANAS \cite{carlucci2019manas}}& 83.62	& 92.67	& 90.11	& 93.45	& 96.09	& 90.34	& 96.09	& 90.84	& 87.88	& 92.47	&  91.24\\
    & \footnotesize{\textpm~1.10} & \footnotesize{\textpm~5.87} & \footnotesize{\textpm~6.46} & \footnotesize{\textpm~6.34} & \footnotesize{\textpm~5.53} & \footnotesize{\textpm~6.24} & \footnotesize{\textpm~4.85} & \footnotesize{\textpm~4.58} & \footnotesize{\textpm~7.46} & \footnotesize{\textpm~8.28} & \footnotesize{\textpm~7.77} \\
    \hline
    \multirow{2}{*}{EWC \cite{kirkpatrick2017overcoming}}& 58.81 & 64.90	& 59.98	& 66.86	& 80.59	& 60.26	& 76.35	& 60.79	& 59.65	& 73.52	& 68.86	\\
    & \footnotesize{\textpm~1.95} & \footnotesize{\textpm~15.56} & \footnotesize{\textpm~17.94} & \footnotesize{\textpm~21.75} & \footnotesize{\textpm~16.92} & \footnotesize{\textpm~20.37} & \footnotesize{\textpm~19.31} & \footnotesize{\textpm~17.17} & \footnotesize{\textpm~19.77} & \footnotesize{\textpm~24.45} & \footnotesize{\textpm~21.26} \\
    \hline
    \multirow{2}{*}{RPS \cite{rajasegaran2019random}}& 18.24	& 61.67	& 67.06	& 74.25	& 83.18	& 70.51	& 82.35	& 83.87	& 76.04	& 76.80	& 71.64	\\
   & \footnotesize{\textpm~8.99} & \footnotesize{\textpm~14.19} & \footnotesize{\textpm~16.31} & \footnotesize{\textpm~14.14} & \footnotesize{\textpm~19.94} & \footnotesize{\textpm~18.84} & \footnotesize{\textpm~9.27} & \footnotesize{\textpm~7.47} & \footnotesize{\textpm~17.10} & \footnotesize{\textpm~23.62} & \footnotesize{\textpm~28.38} \\
    \hline
    \end{tabular}
    \caption{Table compares the validation accuracy of DL models trained with different learning methods on five tasks defined on the CIFAR10 dataset and five tasks defined on the CIFAR100 dataset. Definition of, the tasks are provided in section 4.2.1 in the original paper. For each learning method, five DL models were trained and the mean and standard deviation of validation accuracy for all five models are shown. }
    \label{tab:cifar10_100}
\end{table}
\end{document}